\documentclass[lettersize,journal]{IEEEtran}
\usepackage{amsmath,amsfonts}
\usepackage{algorithmic}
\usepackage{algorithm}
\usepackage{array}
\usepackage[caption=false,font=normalsize,labelfont=sf,textfont=sf]{subfig}
\usepackage{textcomp}
\usepackage{stfloats}
\usepackage{url}
\usepackage{verbatim}
\usepackage{graphicx}
\usepackage{cite}
\usepackage{svg}
\usepackage{booktabs}
\usepackage{multirow}
\usepackage{adjustbox}
\usepackage{tabularx}
\usepackage[hidelinks]{hyperref}
\usepackage{hypcap}
\usepackage{verbatim}
\usepackage{colortbl}
\usepackage{siunitx}
\usepackage{float}
\usepackage{pifont}
\usepackage{makecell}
\usepackage{pifont}
\usepackage{mathtools}

\begin{document}

\title{DGTRSD \& DGTRS‑CLIP: A Dual‑Granularity Remote Sensing Image–Text Dataset and Vision Language Foundation Model for Alignment}

\author{
\begin{center}
\IEEEauthorblockN{Weizhi Chen, Yupeng Deng*, Jin Wei, Jingbo Chen, Jiansheng Chen*, \\
Yuman Feng, Zhihao Xi, Diyou Liu, Kai Li,~\IEEEmembership{Student Member,~IEEE}, Yu Meng}
\end{center}
\thanks{*Corresponding author: Yupeng Deng (dengyp@aircas.ac.cn), Jiansheng Chen (chenjs@aircas.ac.cn). This research was funded by the National Key R\&D Program of China under grant number 2021YFB3900504.}
\thanks{Weizhi Chen, Kai Li are with Aerospace Information Research Institute, Chinese Academy of Sciences, Beijing 100101, China, and also with School of Electronic, Electrical and Communication Engineering, University of Chinese Academy of Sciences, Beijing 100049, China.}
\thanks{Yupeng Deng, Jiansheng Chen, Jingbo Chen, Zhihao Xi, Diyou Liu, Yu Meng are with Aerospace Information Research Institute, Chinese Academy of Sciences, Beijing 100101, China.}
\thanks{Jin Wei is currently working as an engineer in PLA Unit 32021.}
\thanks{Yuman Feng is with the School of Information Network Security, People's Public Security University of China, Beijing 100038, China.}
}

\maketitle

\begin{abstract}
Vision Language Foundation Models based on CLIP architecture for remote sensing primarily rely on short text captions, which often result in incomplete semantic representations. Although longer captions convey richer information, existing models struggle to process them effectively because of limited text‑encoding capacity, and there remains a shortage of resources that align remote sensing images with both short text and long text captions. To address this gap, we introduce DGTRSD, a dual-granularity remote sensing image-text dataset, where each image is paired with both a short text caption and a long text description, providing a solid foundation for dual-granularity semantic modeling. Based on this, we further propose DGTRS-CLIP, a dual-granularity curriculum learning framework that combines short text and long text supervision to achieve dual-granularity semantic alignment. Extensive experiments on four typical zero-shot tasks: long text cross-modal retrieval, short text cross-modal retrieval, image classification, and semantic localization demonstrate that DGTRS-CLIP consistently outperforms existing methods across all tasks. The code has been open-sourced and is available at \url{https://github.com/MitsuiChen14/DGTRS}.
\end{abstract}

\begin{IEEEkeywords}
  Remote Sensing; VLFM; Dual-Granularity; Curriculum Learning; Cross-Modal Alignment.
\end{IEEEkeywords}
\section{Introduction}
\IEEEPARstart{I}{N} recent years, Vision-Language Foundation Models (VLFM) have rapidly advanced in cross-modal semantic alignment and joint representation learning between language and images~\cite{zhang2024vision,du2022survey}. Driven by the growing demand for intelligent analysis in the remote sensing domain, researchers have explored various tasks such as image classification, retrieval, and segmentation using VLFM like CLIP~\cite{radford2021learning}, achieving notable progress~\cite{li2023rs,mall2023remote}. However, remote sensing data are significantly influenced by factors such as scale variations, atmospheric conditions, imaging angles, neighboring interference, and transmission effects~\cite{zhang2025core, wang2024odinmj, ouyang2024multimodal}, which result in substantial differences from natural images. As a result, current research has begun to focus on developing VLFM specifically tailored for the remote sensing domain. Notably, many scholars have attempted to fine-tune models like CLIP with remote sensing-specific image-text pairs, leading to preliminary advancements in tasks such as remote sensing classification and cross-modal retrieval~\cite{liu2024remoteclip,wang2024skyscript,zhang2024rs5m}.
\par
\begin{figure*}[htbp]
  \centering
  \includegraphics[width=\linewidth]{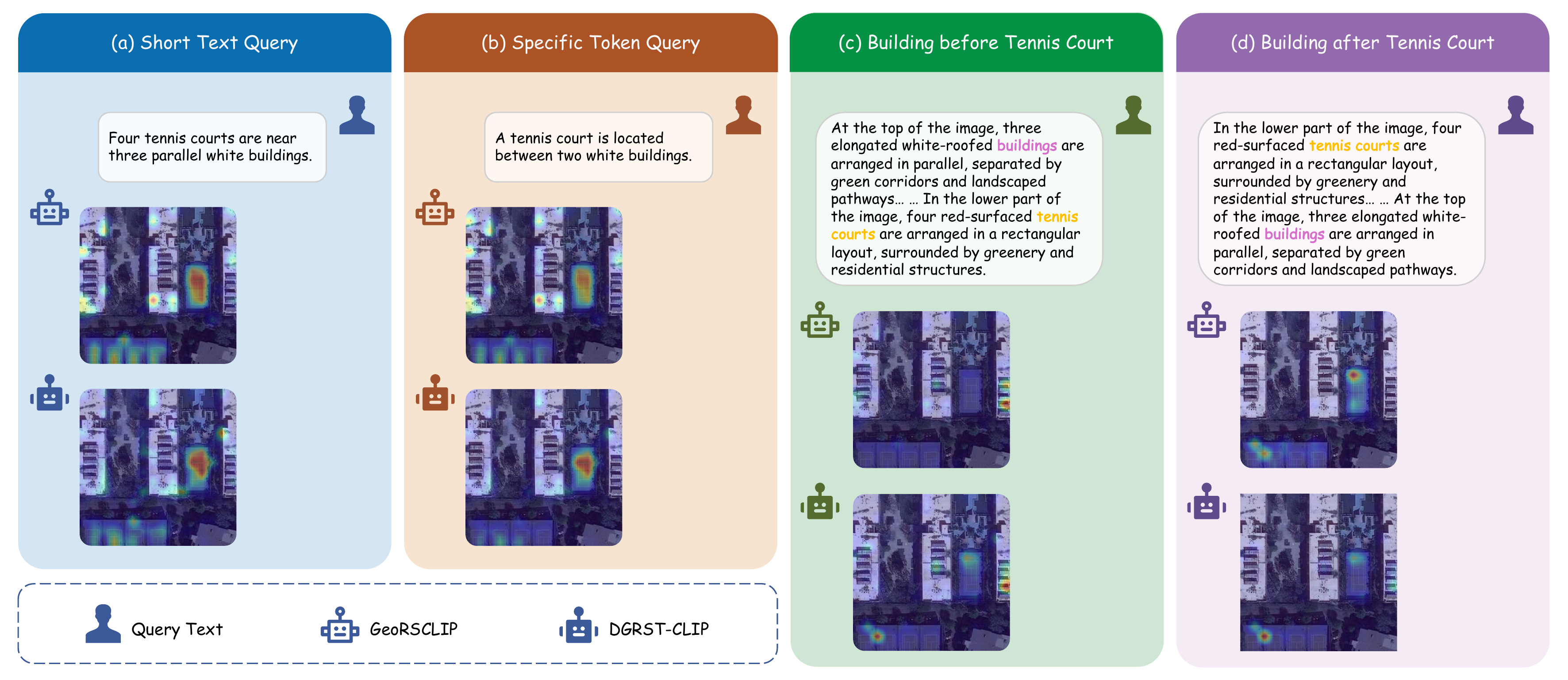}
  \caption{(a) When using a general short text query, our method and GeoRSCLIP exhibit nearly identical visual attention. (b) However, under the specific short text query “A tennis court is located between two white buildings,” GeoRSCLIP focuses on the “tennis court” but neglects its spatial relation to the surrounding “buildings,” indicating insufficient modeling of inter-object spatial relationships. (c) and (d) In the case of long queries with reversed token order (e.g., “buildings” preceding “tennis court”), GeoRSCLIP assigns disproportionately high attention to earlier tokens while largely ignoring the latter, revealing unreasonable attention allocation and limited long text comprehension. In contrast, DGTRS-CLIP consistently captures spatial relationships between objects and maintains balanced attention distribution in long text scenarios, demonstrating superior capability in complex semantic modeling.}
  \label{interpretation}
\end{figure*}
\begin{figure}[htbp]
  \centering
  \includegraphics[width=\linewidth]{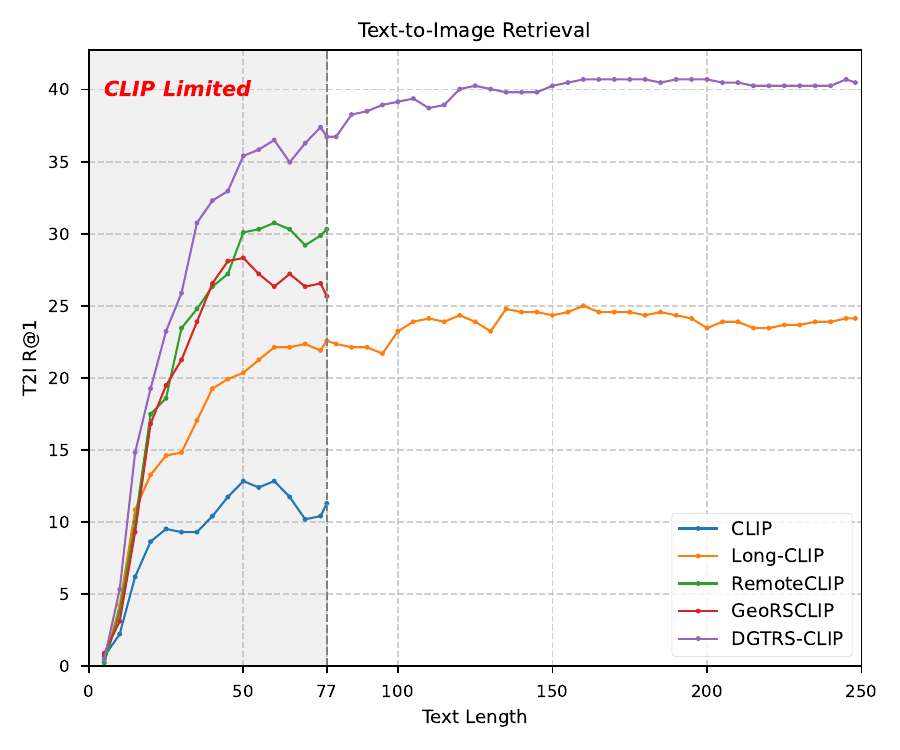}
  \caption{R@1 of different model with varying text lengths. CLIP, RemoteCLIP, and GeoRSCLIP exhibit performance degradation near the maximum encoding length, whereas our model and Long-CLIP, benefiting from joint training with long and short texts, achieve consistent improvements and maintain stable recall.}
  \label{t2i_r1}
\end{figure}
\par
{However, since the training texts are typically short, existing models still struggle with structured semantic understanding, particularly in modeling spatial relationships among objects and in effectively allocating attention. Regarding spatial relationships, as illustrated in Fig.~\ref{interpretation}(b), when querying the image with the description “A tennis court is located between two white buildings”, GeoRSCLIP trained solely on short texts tend to distribute attention across all tennis courts in the scene rather than focusing on the specified one. We attribute this limitation to the lack of spatial relational information in the short-text-based pretraining data.}
{In terms of attention allocation, as shown in Fig.~\ref{interpretation}(c) and (d), GeoRSCLIP consistently attends to the tokens appearing earlier in the sequence, regardless of the actual order of “buildings” and “tennis court,” indicating that its attention is predominantly concentrated on low-index tokens with limited allocation to high-index ones. In contrast, our DGTRS-CLIP effectively addresses both issues, demonstrating improved capability in capturing spatial relationships and distributing attention more appropriately. According to the experimental findings of Zhang et al.~\cite{zhang2024long}, when the token length of text exceeds 20, the improvement of cross-modal retrieval performance in terms of R@1 becomes significantly slower, indicating that the effective encoding length of CLIP does not surpass 20 tokens. To further investigate this issue, we conducted retrieval experiments on several representative CLIP-based architectures in both general and remote sensing domains. As shown in Fig.~\ref{t2i_r1}, the retrieval accuracy of CLIP increases slowly once the text length reaches 20 tokens and even declines when approaching the maximum token limit. Compared with CLIP, RemoteCLIP and GeoRSCLIP exhibit relatively longer effective retrieval lengths, as they are pre-trained on remote sensing image-text datasets with longer textual descriptions. Similar to CLIP, their retrieval accuracy degrades as the input length approaches the encoder limit. We argue that the root cause lies in CLIP's primary pretraining on short texts, leading to uneven token training across longer sequences.}
\par
{But relying solely on long-text training also presents drawbacks. Since the CLIP text encoder is inherently biased toward low-index tokens (particularly the first 20) due to pretraining, and its native input length is restricted to 77 tokens, directly extending the sequence length can exacerbate positional biases in the Transformer architecture. As a result, the model’s attention becomes excessively concentrated on the initial tokens, while newly added high-index tokens—insufficiently trained during pretraining—yield weak or even noisy semantic representations. Consequently, critical information in the latter part of long texts (e.g., spatial relations among objects) is severely underrepresented, while redundant information and ineffective representations of added tokens accumulate, further diluting the core semantics. This ultimately impairs the model’s ability to focus on key information during cross-modal matching and leads to a decline in rapid retrieval performance.}
\par
Consequently, constructing a VLFM capable of balancing coarse-grained and fine-grained semantic expression to support multi-level semantic modeling and spatial relation understanding, including tasks such as cross-modal retrieval and visual grounding, has become an urgent research problem. To achieve this objective, two key challenges remain. First, data resources are insufficient with respect to semantic granularity. Commonly used remote sensing image–text datasets, including TextRS~\cite{abdullah2020textrs}, RSICD~\cite{lu2017exploring}, RSITMD~\cite{yuan2022exploring}, UCMerced~\cite{yang2010bag}, and Sydney-Captions~\cite{qu2016deep}, mainly contain short captions that cover coarse-grained semantics, whereas datasets such as VersaD~\cite{pang2025vhm} primarily offer long text captions focusing on fine-grained semantics. Hence, current datasets lack unified multi-granularity annotations, which limits the ability of models to align and model information at different semantic levels. Second, training strategies are restrictive. Existing CLIP-based approaches in the remote sensing domain have long relied on texts of a single granularity, making it difficult for the models to balance complex semantic understanding and succinct expression, and thereby impeding the synergistic fusion and unified modeling of multi-granularity semantics.
\par
To address the aforementioned challenges in both data and training strategies, this study adopts a two-pronged approach, focusing on data construction and model design. To mitigate the limitations of semantic granularity in existing datasets, DGTRSD has been developed using annotation methods based on large language models (LLMs) and multimodal large language models (MLLMs). Comprising approximately 1,762,131 image–text pairs, DGTRSD provides, for the first time, both short and long image captions for each image, thus establishing a robust foundation for multi-granularity semantic modeling.
\par
Based on DGTRSD, a dual-granularity curriculum learning transfer framework has been introduced to overcome the limited transferability of generic pretrained models to the remote sensing domain and their inadequate capacity for long text modeling. This framework integrates Knowledge Preserved Stretching (KPS)~\cite{zhang2024long}, which effectively extends the encodable text length of the CLIP model and enhances its ability to represent long textual information. In conjunction with a dual-granularity curriculum learning strategy, the framework dynamically balances the roles of long and short texts in aligning with remote sensing imagery, thereby facilitating the fusion and unified modeling of multi-granularity semantic information.
\par
To comprehensively evaluate the effectiveness and generalization capability of the proposed method, four typical zero-shot tasks have been established for remote sensing vision language understanding: long text cross-modal retrieval (LTCR), short text cross-modal retrieval (STCR), image classification (IC), and semantic localization (SeLo)~\cite{yuan2022learning}.
\par
The principal contributions of this paper are summarized as follows:
(1) A novel remote sensing image-text dataset, DGTRSD, is developed that provides dual-granularity annotations, including both short and long textual descriptions for each image.
(2) A CLIP-based dual-granularity curriculum learning framework is proposed, effectively mitigating the limited transferability of general vision language models to remote sensing applications and overcoming their constraints in long text modeling.
(3) Four zero-shot remote sensing tasks are designed: LTCR, STCR, IC and SeLo. Experimental results show that the proposed DGTRS-CLIP model consistently outperforms existing methods across all tasks, demonstrating superior semantic expressiveness and generalization, particularly in complex semantic-understanding scenarios and multi-object remote sensing image analysis.
\section{Related work}
\subsection{VLFM for Remote Sensing}
The development of foundational theories in multimodal representation learning has laid an important foundation for VLFM research. By jointly encoding images and text, VLFM aims to construct a unified model architecture that efficiently integrates and understands information from both modalities. In recent years, with the introduction of large-scale pretrained models such as CLIP~\cite{radford2021learning}, VLFM applications have made breakthrough progress. These models leverage large-scale datasets for pretraining and demonstrate strong transfer learning abilities across various downstream tasks. In the remote sensing domain, images often have high resolution and diverse land cover types, while textual descriptions contain rich semantic information related to these objects. VLFM bridges the semantic gap between images and text, opening up new possibilities for intelligent remote sensing data processing.

From a methodological perspective, VLFM research can be broadly divided into two main paradigms: contrastive learning and generative learning~\cite{xiao2024foundation}. Contrastive learning-based VLFM enhances the semantic alignment between modalities by maximizing the similarity between images and text. Specifically, the model uses an image encoder and a text encoder to map images and text to a shared embedding space, using contrastive losses such as Triplet Loss~\cite{schroff2015facenet} or InfoNCE Loss~\cite{oord2018representation} to pull together correctly matched image-text pairs and push apart mismatched pairs. In contrast, generative learning-based VLFM focuses on generative tasks between images and text. For example, Muhtar et al.~\cite{muhtar2024lhrs} combined multi-level vision language alignment strategies with curriculum learning methods to generate contextually appropriate text descriptions and perform detailed reasoning.

CLIP~\cite{radford2021learning} has used contrastive learning on 300 million image-text pairs to map images and text to a unified embedding space successfully. This has led to the development of several model variants, such as SigLIP~\cite{zhai2023sigmoid}, SLIP~\cite{mu2022slip}, Structure-CLIP~\cite{huang2024structure}, MaskCLIP~\cite{zhou2022extract}, and CLIPPO~\cite{tschannen2023clippo}. These models have achieved outstanding performance in cross-modal retrieval and zero-shot tasks. However, since CLIP does not support tasks such as object detection or visual grounding, researchers often need to design additional modules for integration. For example, Region-CLIP~\cite{zhong2022regionclip} improves the pretrained CLIP model with an R-CNN style detector, enabling strong performance in open-vocabulary object detection tasks. CLIP-VG~\cite{xiao2023clip} implements visual grounding through a simple Transformer fusion module, and RISCLIP~\cite{kim2023extending} treats CLIP image and text encoders as backbone networks, introducing adaptive architectures for reference segmentation tasks.

Transferring CLIP from the general domain to the remote sensing domain presents a key challenge: creating suitable remote sensing datasets and building baseline models through supervised or task-specific finetuning. Studies such as RemoteCLIP~\cite{liu2024remoteclip}, SkyCLIP~\cite{wang2024skyscript}, and GeoRSCLIP~\cite{zhang2024rs5m} have finetuned on customized image caption datasets while retaining the original CLIP architecture and have been evaluated on downstream tasks such as cross-modal retrieval and zero-shot classification. GeoRSCLIP achieved state-of-the-art semantic localization performance on the AIR-SLT~\cite{yuan2022learning} dataset. However, most of these studies focus on aligning short text captions with remote sensing images, primarily emphasizing local or salient features, while often lacking a deeper understanding of global and fine-grained information. Meanwhile, some studies have focused on aligning remote sensing images with geographic coordinates. For example, GeoCLIP~\cite{vivanco2023geoclip} and SatCLIP~\cite{klemmer2023satclip} have finetuned the geographic coordinate encoding and image encoding in a contrastive manner similar to CLIP, successfully achieving tasks like Image-GPS retrieval.

\subsection{Dual-Granularity Text Alignment in VLFM}
In traditional cross-modal retrieval systems, long texts are typically treated as a collection of short texts directly matched with images. However, long texts often contain rich semantic information, such as multiple land cover categories, spatial relationships, and temporal context. This information must be fully extracted and organized to accurately describe and match the corresponding images. Contrastive learning models such as CLIP are not designed to focus on long text processing. Their text encoders are usually limited to 77 tokens, which challenges aligning long texts with images.

To the best of our knowledge, existing VLFM have not yet fully addressed the challenges of understanding long texts in the remote sensing domain. However, some relevant explorations have gradually emerged in the general domain. The early DreamLIP~\cite{zheng2024dreamlip} research randomly sampled sub-captions from detailed long text captions for training. While this approach somewhat improved the model's understanding of long texts, it did not fully utilize the complete long text information. Long-CLIP~\cite{zhang2024long} expanded the text encoder's effective token length based on the original CLIP through methods like Knowledge Preserved Stretching (KPS) and Primary Component Matching (PCM). Additionally, it treated the first sentence of a long text (typically a summary or overview) as a short text input, maintaining good adaptation to short texts while enhancing long text comprehension. MATE~\cite{jang2024mate} replaced the text encoder in VLM with a pretrained encoder based on LLMs and used multi-stage training to align the embeddings of images and LLMs, overcoming VLM's limitations in processing long captions or documents. LoTLIP~\cite{wu2024lotlip} re-annotated the data with long captions and incorporated corner annotations to integrate different types of textual information, thereby enhancing the model's ability to understand long texts while maintaining its performance on short text tasks. TULIP~\cite{najdenkoska2024tulip} extended CLIP's token processing capacity by using relative position distillation and relative position expansion, unlocking new possibilities for efficiently processing long texts.

Although the aforementioned methods have made certain progress in long text modeling within the domain of natural images, most existing studies focus only on aligning long or short texts with images in general scenarios. They have yet to fully address the unique challenges of remote sensing image-text tasks, such as multi-object, multi-scale, and spatial relationship modeling. Moreover, they generally lack mechanisms for joint modeling of short and long texts tailored to the characteristics of remote sensing data. Therefore, designing vision language models capable of unified modeling of short and long text semantic information has become a critical research direction for advancing remote sensing image-text understanding and improving performance in downstream tasks.
\section{DATASET CONSTRUCTION}
In this study, a remote sensing image–text dataset named DGTRSD was constructed to provide rich and diverse semantic information for image–text understanding tasks. DGTRSD comprises paired long text and short text captions for every image. Its images and captions were drawn from open-source remote sensing datasets, including the long text dataset VersaD~\cite{pang2025vhm}, the short text datasets RS5M~\cite{zhang2024rs5m}, UCMerced~\cite{yang2010bag}, Det-10, and Seg-4~\cite{liu2024remoteclip}, as well as the high-resolution, object-level annotation dataset OpenLandMap, which was curated in-house. To ensure that each remote sensing image is accompanied by both a short text caption and a long text caption, the dataset was compiled through three sequential stages: (1) image collection, (2) text supplementation, and (3) data evaluation{, (4) dataset statistics}.
\begin{figure*}[htbp]
  \centering
  \includegraphics[width=\linewidth]{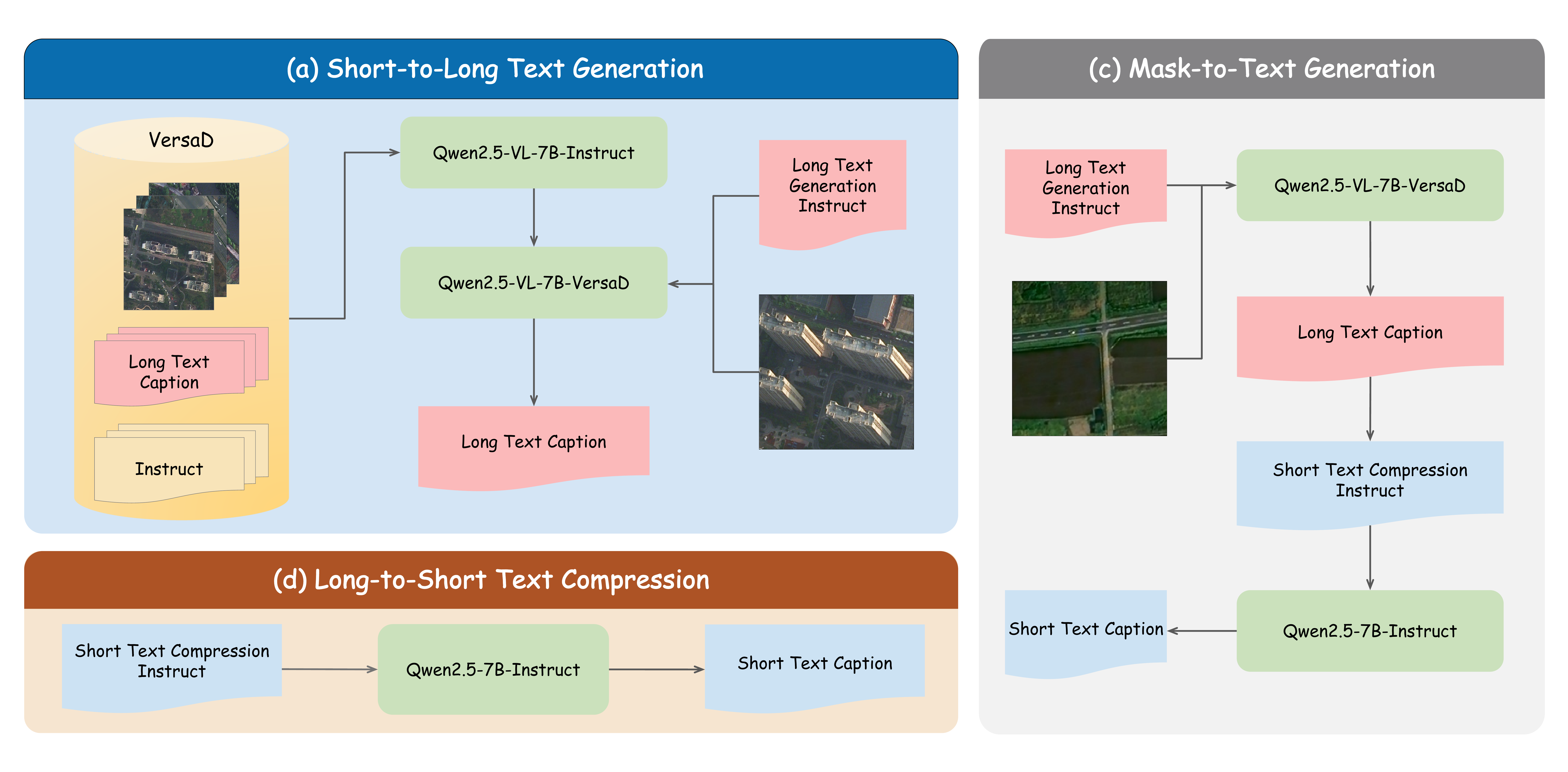}
  \caption{{Overview of the Text Supplementation pipeline. (a) Short-to-Long Text Generation refers to the process of generating long texts from datasets containing short texts, which is primarily applied to RS5M, Det-10, Seg-4, and UCMerced. First, by leveraging the Instruct-image-caption triplets from VersaD, we finetune the Vision Encoder and Vision–Language Connector modules of Qwen2.5-VL-7B-Instruct using LoRA sft. The finetuned model is named Qwen2.5-VL-7B-VersaD. Based on this model, we generate long captions for the target images following the instructions shown in Fig.~\ref{inst} (a). (b) Long-to-Short Text Compression involves generating short texts from datasets with long text annotations, mainly applied to VersaD. Different from (a), we directly integrate the long captions with compression instructions (Fig.~\ref{inst} (b)) as input to Qwen2.5-7B-Instruct, and obtain concise short texts. (c) Mask-to-Text Generation denotes the pipeline for generating both long and short texts based on mask annotations, primarily applied to OpenLandMap. First, long captions are generated following the method in (a), and then short texts are obtained following the method in (b). The specific instructions are illustrated in Fig.~\ref{inst} (c).}}
  \label{fig_2}
\end{figure*}
\begin{figure*}[htbp]
  \centering
  \includegraphics[width=\linewidth]{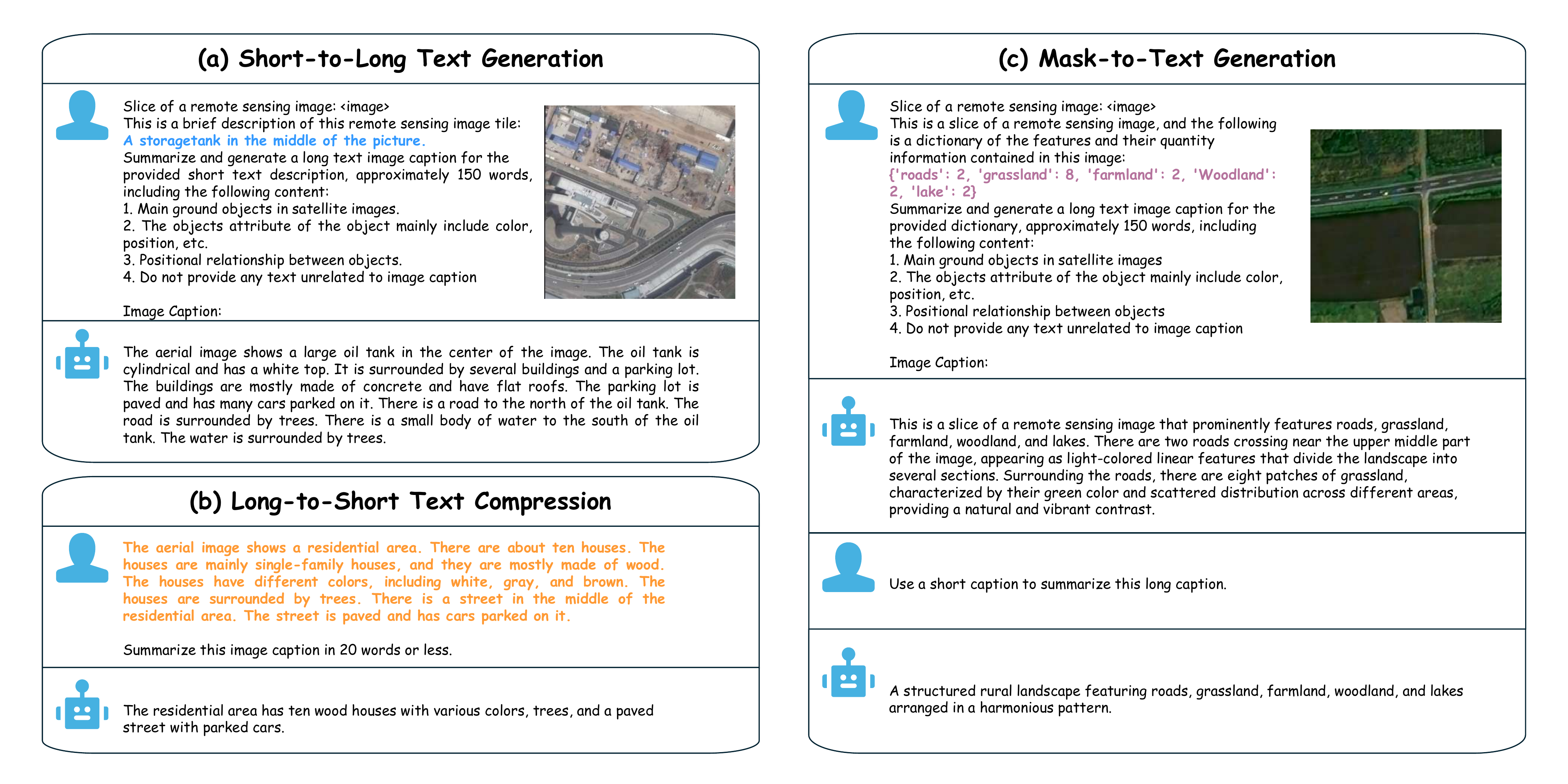}
  \caption{{DGTRSD generation prompts. In the prompts, blue text denotes the short text caption of the current image, orange text indicates the long text caption prompt, and purple text represents the masked prompt of the current image.}}
  \label{inst}
\end{figure*}
\subsection{Image Collection}
DGTRSD was constructed from multiple publicly available remote sensing image–text corpora, specifically the long text dataset VersaD and the short text datasets RS5M, UCMerced, Det-10, and Seg-4. In addition, a laboratory-curated, high-precision, object-level annotation set, OpenLandMap, was incorporated to further diversify the data sources.
During collection, approximately 14 million remote sensing images were retrieved from VersaD, covering several public repositories—including MillionAID~\cite{long2021creating}, CrowdAI~\cite{mohanty2020deep}, fMoW~\cite{christie2018functional}, CVACT~\cite{liu2019lending}, CVUSA~\cite{zhai2017predicting}, and LoveDA~\cite{wang2021loveda}—with spatial resolutions ranging from 0.08 m to 153 m per pixel. For RS5M, the MillionAID and BigEarthNet~\cite{sumbul2019bigearthnet} subsets were selected, noting that the MillionAID portion partially overlaps with that in VersaD. Additional imagery was gathered from the RemoteCLIP training set Det-10, which aggregates ten detection datasets (AUAIR~\cite{vujasinovic2020integration}, CARPK~\cite{hsieh2017drone}, DIOR~\cite{li2020object}, DOTA~\cite{xia2018dota}, HRRSD~\cite{zhang2019hierarchical}, HRSC~\cite{liu2017high}, LEVIR~\cite{chen2020spatial}, RSOD~\cite{sun2022rsod}, Stanford~\cite{robicquet2016learning}, and VisDrone~\cite{zhu2018vision}). Further data were sourced from Seg-4, comprising four segmentation datasets (iSAID~\cite{waqas2019isaid}, LoveDA~\cite{wang2021loveda}, Potsdam~\cite{2DSemanticLabel}, and Vaihingen~\cite{2DSemanticLabeling}), as well as from the UCMerced archive. Finally, 27,533 additional remote sensing image tiles with 0.5 m spatial resolution and high-precision object-level labels were incorporated from the in-house OpenLandMap collection.
\subsection{Text Supplementation}
To systematically construct remote sensing image–text pairs with dual-granularity descriptions, a text-supplementation scheme comprising four subtasks was devised. A dedicated strategy was adopted for each type of data sample, as detailed below. The detailed instruction is provided in {Fig.~\ref{inst}}.
\par
\textbf{1) Existing Caption Consolidation.} VersaD and RS5M respectively provide long text and short text captions for the MillionAID images. These captions were directly merged, yielding a foundational subset in which every image is associated with both a long and a short caption.
\par
{\textbf{2) Short-to-Long Text Generation.} In the RS5M BigEarthNet subset, UCMerced, Det-10, and Seg-4 datasets, each image is annotated with a single short caption. To supplement the missing long descriptions, we designed an MLLM-based short-text-guided long-text caption generation approach, as illustrated in Fig.~\ref{fig_2}(a). Specifically, through prompt engineering, the MLLM was instructed to generate long captions conditioned on the provided short text, with explicit emphasis on \textbf{(1) the primary ground objects in the image}, \textbf{(2) the attributes of these objects}, and \textbf{(3) the spatial relationships among them}, as exemplified in Fig.~\ref{inst}(a). However, experiments across several MLLMs revealed that, due to the long-tailed distribution of remote sensing scenes, the generated captions were often suboptimal in quality. To address this issue, 10\% of the VersaD dataset was selected for finetuning. Each sample consists of an instruction fed to Gemini-Vision~\cite{team2023gemini}, paired with the resulting image caption and its corresponding remote sensing image. According to the dataset provider, VersaD achieved an overall accuracy of 82.3\% in a human sampling evaluation, suggesting that it provides sufficiently reliable captions for finetuning.}
\par
{Based on VersaD, we conducted LoRA~\cite{hu2022lora} finetuning on several candidate backbone models, including Qwen2.5-VL-7B-Instruct~\cite{xu2025qwen2}, InternVL2.5-8B, LLaVA-OneVision-Qwen2-7B-OV-HF, and GLM-4V-9B. Considering the discrepancy between VersaD instructions and synthetic data instructions, as well as the risk of catastrophic forgetting when finetuning LLM modules, we restricted finetuning to the Vision Encoder and Vision–Language Connector modules, with the rank uniformly set to 16. Following a rapid manual inspection of small batches of captions generated by the finetuned models, Qwen2.5-VL-7B-Instruct was selected and renamed as Qwen2.5-VL-7B-VersaD10. This model was subsequently employed for long caption generation using the prompts shown in Fig.~\ref{inst}(a).}
\par
{\textbf{3) Long-to-Short Text Compression.} To avoid visual interference from remote sensing images during textual modeling, the long-to-short text generation was designed as a pure text summarization task. Since this task does not involve remote sensing visual knowledge, the choice of LLM backbone is less critical. As shown in Fig.~\ref{fig_2}(b), we used Qwen2.5-7B-Instruct\cite{yang2024qwen2} to compress long captions from the VersaD CrowdAI, fMoW, CVACT, CVUSA, and LoveDA subsets into concise and accurate short descriptions. An example of the prompt design is provided in Fig.~\ref{inst}(b).}
\par
{\textbf{4) Mask-to-Text Generation.} The object-level masks manually annotated in OpenLandMap were employed as semantic anchors. First, object labels and their counts were used to replace short-text prompts, enabling the generation of detailed long captions through the Short-to-Long Text Generation process, with prompt details illustrated in Fig.~\ref{inst}(c). Subsequently, using the same prompt design as Long-to-Short Text Compression, the corresponding short captions were distilled from the generated long captions, thereby ensuring semantic consistency and precision across both granularities. The overall process is illustrated in Fig.~\ref{fig_2}(c).}
\subsection{Data Evaluation}
To guarantee data reliability and to avoid biases introduced by source models, the generated image–text pairs were subjected to a comprehensive quality assessment using an independent vision–language model, LLaVA-OneVision-Qwen2-7B-OV-HF~\cite{li2024llava}. The evaluation focused on three dimensions:
\par
\textbf{1) Image text consistency:} whether the visual content is faithfully reflected in the textual description.
\par
\textbf{2) Long text fidelity:} whether the long caption provides a detailed depiction of objects, attributes, and spatial relationships.
\par
\textbf{3) Short text adequacy:} whether the short caption concisely and accurately summarizes the principal objects in the image.
\par
A rigorous scoring criteria was established on a 0–5 score. Samples receiving a score of 3 or higher were deemed valid, whereas those below 3 were discarded. The results of the evaluation are shown in the {TABLE~\ref{tab1}}. After this filtering process, the resulting DGTRSD was both reliable and semantically rich, providing a robust, high-quality foundation for remote sensing image–text understanding, cross-modal learning, and intelligent remote sensing analysis.
\par
\begin{figure}[htbp]
  \centering
  \includegraphics[width=\linewidth]{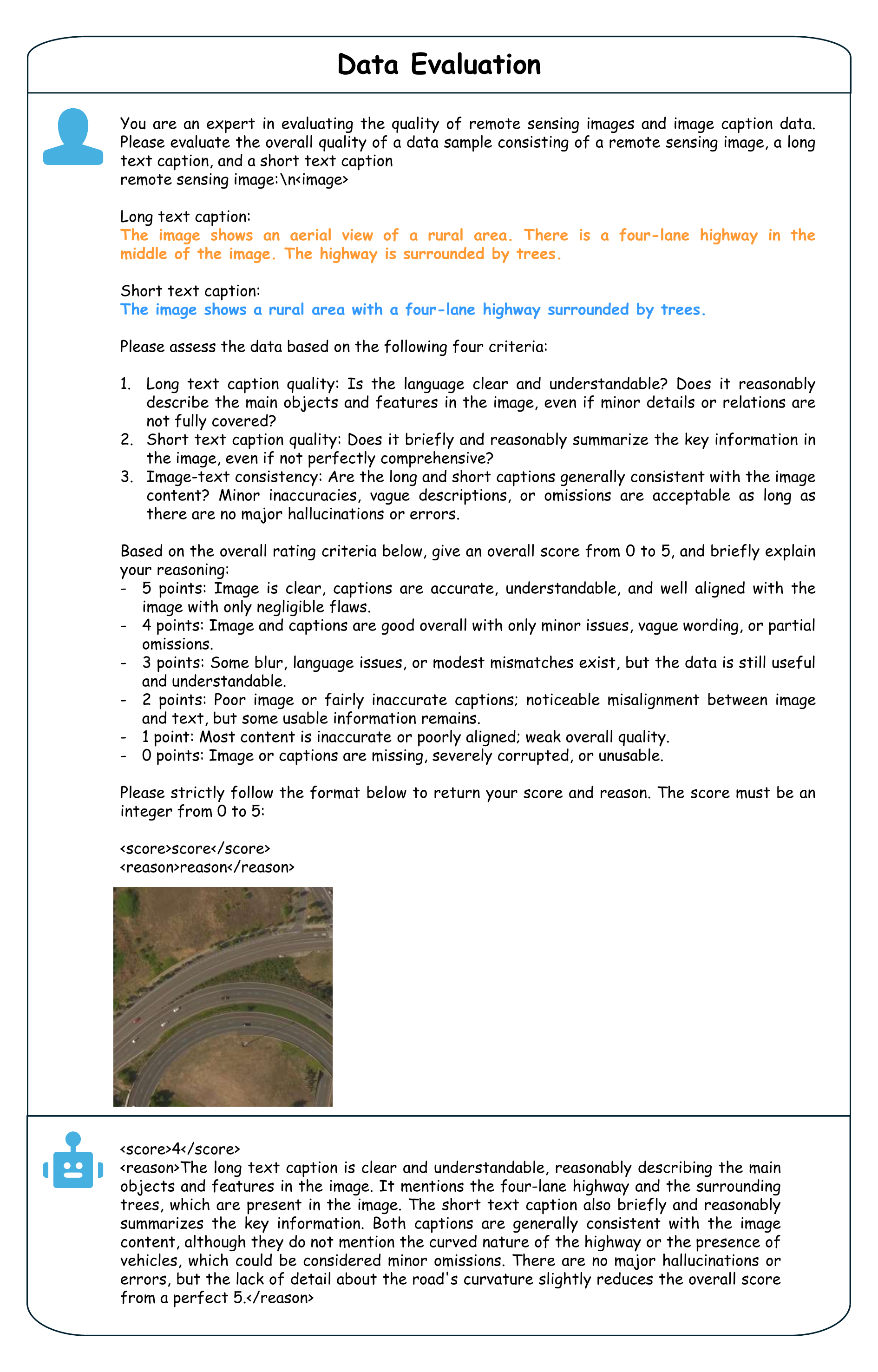}
  \caption{{DGTRSD Evaluation prompt. In the instruction, blue text denotes the short text caption of the current image, orange text indicates the long text caption prompt.}}
  \label{eval}
\end{figure}
To further enhance the interpretability and transparency of the evaluation, the employed MLLM was prompted not only to assign numerical scores but also to generate brief justifications for each judgment. This strategy enables human reviewers to better understand the rationale behind model decisions, facilitates error analysis, and helps identify systematic failure cases in model reasoning.Moreover, generating justifications encourages the model to internalize and align with the evaluation criteria more effectively, leading to more accurate and consistent scoring. {The detailed evaluation prompts are illustrated in Fig.~\ref{eval}.}
\begin{table*}[htbp]
\caption{{The scoring criteria and score distribution of DGTRS are illustrated. Samples that either fail to generate valid reasoning results or produce outputs inconsistent with the evaluation template are assigned a score of 0.}}
\centering
\renewcommand{\arraystretch}{1.2}
\begin{tabular*}{\textwidth}{
    @{\extracolsep{\fill}}
    >{\centering\arraybackslash}m{1.2cm}
    >{\raggedright\arraybackslash}m{12cm}
    >{\raggedleft\arraybackslash}m{2.0cm}
    @{}
}
\toprule
\textbf{Score} & \textbf{Criteria} & \textbf{Count} \\
\midrule
0 & Image or captions are missing, severely corrupted, or unusable. & \num{209} \\
1 & Most content is inaccurate or poorly aligned; weak overall quality. & \num{23424} \\
2 & Poor image or fairly inaccurate captions, noticeable misalignment between image and text, but some usable information remains. & \num{13191} \\
3 & Some blur, language issues, or modest mismatches exist, but the data is still useful and understandable. & \num{516588} \\
4 & Image and captions are good overall with only minor issues, vague wording, or partial omissions. & \num{1245460} \\
5 & Image is clear, captions are accurate, understandable, and well aligned with the image with only negligible flaws. & \num{83} \\
\bottomrule
\end{tabular*}
\label{tab1}
\end{table*}
\begin{table*}[h]
\caption{Statistics of DGTRS}
\centering
\renewcommand{\arraystretch}{1.2}
\begin{tabular}{lrrrrr}
\toprule
\textbf{Sub-dataset} & \textbf{Pre-filter Count} & \textbf{Post-filter Count} & \textbf{Retention Rate (\%)} & \textbf{Avg. Long Text Tokens} & \textbf{Avg. Short Text Tokens} \\
\midrule
MillionAID             & 920,057   & 835,801   & 90.84   & 77.96  & 20.51 \\
BigEarthNet            & 327,166   & 280,143   & 85.63   & 91.84  & 66.82 \\
UCMerced               & 1,886     & 1,865     & 98.89   & 97.61  & 13.15 \\
Det-10                 & 110,794   & 108,806   & 98.21   & 84.35  & 12.76 \\
Seg-4                  & 41,171    & 39,235    & 95.30   & 82.02  & 14.65 \\
CrowdAI                & 276,344   & 276,019   & 99.88   & 91.58  & 36.72 \\
CVACT                  & 44,416    & 44,289    & 99.71   & 82.45  & 38.90 \\
CVUSA                  & 44,416    & 44,267    & 99.66   & 83.89  & 37.95 \\
LoveDA                 & 23,948    & 23,606    & 98.57   & 83.08  & 39.21 \\
fMoW-Big (center bbox) & 28,528    & 28,499    & 99.90   & 80.93  & 32.44 \\
fMoW-Small             & 52,696    & 52,662    & 99.94   & 83.16  & 32.16 \\
OpenLandMap            & 27,533    & 26,939    & 97.84   & 93.71  & 39.45 \\
\midrule
DGTRS                  & 1,898,955 & 1,762,131 & 92.79   & 86.05  & 32.06 \\
\bottomrule
\end{tabular}
\label{tab2}
\end{table*}

\subsection{Dataset Statistics}
\begin{figure*}[htbp]
  \centering
  \includegraphics[width=\linewidth]{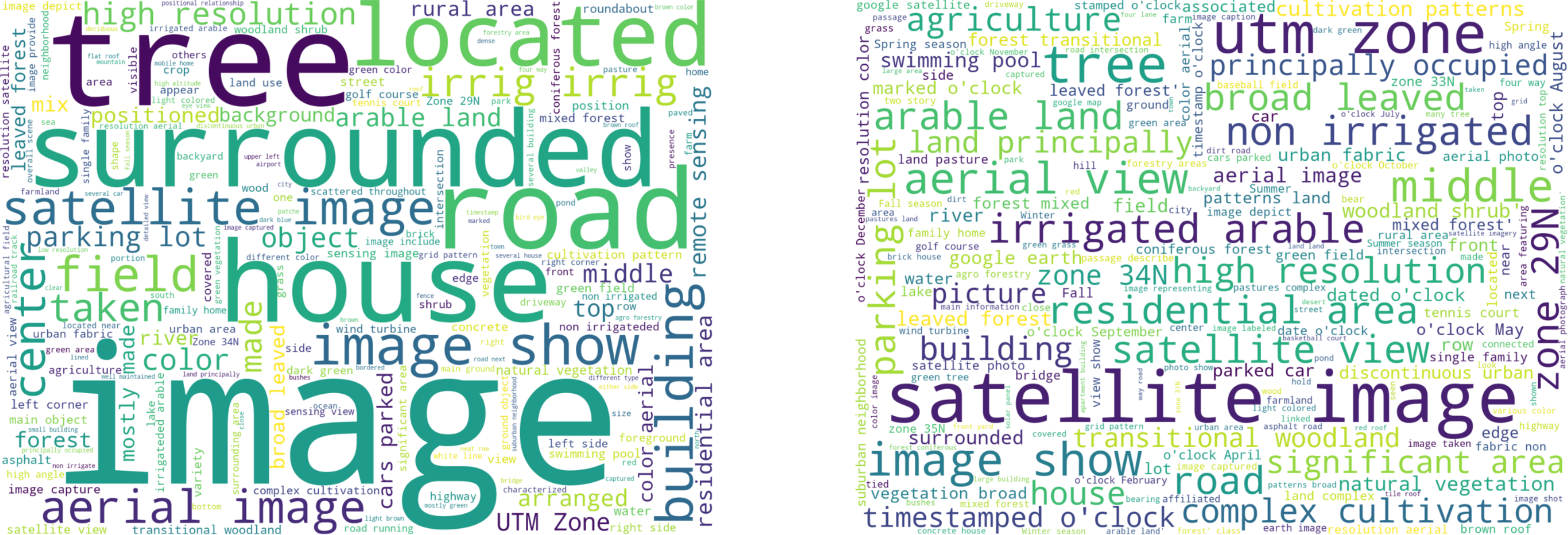}
  \caption{Word clouds of long text (left) and short text (right) image captions. Compared to short captions, long captions contain more spatial relational terms such as \textbf{\textit{located}} and \textbf{\textit{surrounded}}.}
  \label{fig_3}
\end{figure*}
Following the evaluation process, the statistical details of each sub-dataset are presented in Table~\ref{tab2}. The results show that, except for BigEarthNet, all sub-datasets within DGTRS achieve a retention rate above 90\%. Furthermore, the average token length of long text captions in DGTRS reaches 86.05, which is significantly higher than that of short captions (32.06). Fig~\ref{fig_3} illustrates the word clouds of long text and short text image captions, respectively. The comparison between word clouds shows that long text captions contain richer semantic and spatial information, characterized by frequent use of relational terms such as "located" and "surrounded". In contrast, short text captions are more label-like and lack detailed spatial context, indicating their limited capability in modeling complex scene understanding.
\section{DGTRS-CLIP}
To address the dual challenges posed by the long-tail distribution bias inherent in general vision language models and the modality compatibility constraints of the CLIP ~\cite{radford2021learning}architecture in the task of aligning remote sensing image with dual-granularity text semantics, a multi‑stage pretraining framework is proposed in this study. This framework consists of three key components: (1) the backbone network of the CLIP architecture, (2) Knowledge Preserved Stretching (KPS), which extends the text encoding length of the CLIP model, and (3) a dual-granularity curriculum learning mechanism designed to dynamically adjust supervision signals for long text and short text modeling. Through the effective integration of these modules, the proposed framework not only preserves the model aptitude for comprehending short texts but also substantially reinforces its capacity to model long text.
\subsection{CLIP Architecture}
The CLIP proposed by OpenAI adopts a typical two-tower architecture. In this architecture, images and texts are processed through two independent encoders: one image encoder and one text encoder. The image encoder is typically a convolutional neural network (e.g., ResNet~\cite{he2016deep}) or a vision transformer (ViT)~\cite{dosovitskiy2020image}, while the text encoder is usually a Transformer-based model. These two encoders map the images and texts into a shared embedding space. Within this embedding space, CLIP optimizes the model using contrastive learning, ensuring that related image-text pairs are pulled closer, while unrelated pairs are pushed apart.
\par
To implement this contrastive learning, CLIP employs a shared InfoNCE loss function. The goal of this loss function is to maximize the similarity between correctly paired images and texts, while minimizing the similarity between mismatched pairs. For a batch of N samples, the mathematical expression of the InfoNCE loss function is as follows:
\begin{align}
\mathcal{L}_{\text{InfoNCE}} = & -\frac{1}{2N} \sum_{i=1}^N \left[ \log \frac{\exp \left( \text{sim}(v_i, t_i) / \tau \right)} {\sum_{j=1}^{N} \exp \left( \text{sim}(v_i, t_j) / \tau \right)} \right. \label{eq:infoNCE} \\
& + \left. \log \frac{\exp \left( \text{sim}(t_i, v_i) / \tau \right)}{\sum_{j=1}^{N} \exp \left( \text{sim}(t_i, v_j) / \tau \right)} \right] \notag 
\end{align}
where $v_{i}$ represents the encoded feature vector of the $i-th$ image, $t_{i}$ represents the encoded feature vector of the $i-th$ text, and $\tau$ is the temperature parameter used to control the smoothness of the distribution. The function $sim(v_{i}, t_{i})$ represents the cosine similarity between the image and text, which is computed as:
\begin{equation}
  \text{sim}( v_i, t_i ) = \frac{v_i^{\top} t_i}{\lVert v_i \rVert \lVert t_i \rVert}
\end{equation}
\par
With this strategy, negative samples are not explicitly constructed, as in a batch of $N$ samples, all pairs except the diagonal positive pairs are considered negative samples. Moreover, as the batch size $N$ increases, the number of negative samples grows linearly, which contributes to improving the contrastive learning effectiveness.
\subsection{Knowledge Preserved Stretching}
The Knowledge Preserved Stretching (KPS) module, first introduced by Zhang et al.~\cite{zhang2024long}, was devised to overcome the input-length limitation of the CLIP model. KPS achieves length extension by retaining the first 20 well-trained position embeddings (PEs) and applying large-scale interpolation to the remaining PEs, thereby minimizing disturbance to the original positional representations. Specifically, when the position index $\text{p} \le \theta$, the original position embedding is preserved; when $\text{p} >\theta $, a linear interpolation is performed between the two nearest original position embeddings. The procedure is formally expressed as follows:
\begin{equation}
\mathrm{PE}^{\!*}(p) =
\left\{
\begin{array}{@{}l@{\quad}r@{}}
\mathrm{PE}(p), & 0 \le p \le \theta, \\[4pt]
(1 - \omega)\,\mathrm{PE}\!\left( \left\lfloor \dfrac{p}{\lambda} \right\rfloor \right)
+ \omega\,\mathrm{PE}\!\left( \left\lceil \dfrac{p}{\lambda} \right\rceil \right), & p > \theta.
\end{array}
\right.
\label{eq:pe_interp}
\end{equation}
In the above formulation, the interpolation weight is controlled by $ \omega$, and $\lambda$ represents the stretching ratio. The maximum extended input length, denoted as $L_{\max}$, is computed as follows:
\begin{equation}
L_{\max}=\theta +\lambda \left( L_{orign}-\theta \right) 
\label{eq4}
\end{equation}
\par
In this context, the maximum input length of the original CLIP model is denoted as $L_{\text{origin}} = 77$. According to the experimental analysis conducted by Zhang et al.~\cite{zhang2024long}, the optimal threshold is set to $\theta = 20$. As a result, after integrating KPS, the maximum input length supported by the CLIP text encoder is extended from 77 to 248, denoted as $L_{\text{max}}$. This approach not only leverages the well-trained low-position embeddings effectively but also adjusts the higher-position embeddings in a controlled manner, thereby enabling efficient encoding of long textual sequences.
\subsection{Dual-Granularity Curriculum Learning}
To address the challenges associated with the transfer and integration of dual-granularity textual information during model finetuning, this study proposes a Dual-Granularity Curriculum Learning (DGCL) strategy. By dynamically adjusting the loss weight of long text supervision signals, DGCL facilitates effective knowledge transfer from long text to short text representations and promotes joint optimization. Specifically, during the finetuning stage, the total loss is formulated as the weighted sum of losses corresponding to long text and short text supervision. The calculation is expressed as follows:
\begin{equation}
\mathcal{L}_{\text{total}}=\alpha \cdot \mathcal{L}_{\text{long}}+\left( 1-\alpha \right) \cdot \mathcal{L}_{\text{short}}
\end{equation}
where $\mathcal{L}_{\text{long}}$ and $\mathcal{L}_{\text{short}}$ represent the losses computed between the image and long text, and the image and short text, respectively. The coefficient $\alpha$ denotes the weight assigned to the long text loss. By dynamically modulating 
$\alpha$ throughout training, DGCL assigns appropriate optimization priorities to long and short textual inputs at different stages, thereby enhancing the model unified representation capability and generalization across multi-granularity expressions.
\par
{Since CLIP was originally designed for short text modeling, our framework incorporates KPS and DGCL to address the limitations in long text processing. KPS extends the maximum input length of the text encoder and enables effective modeling of semantic dependencies among long text tokens, thereby providing the foundation for DGCL. In the warm-up phase, KPS allows the model to fully leverage long text supervision to capture fine-grained attributes and spatial relationships. During the decay phase, DGCL gradually decreases the weight of long text supervision while strengthening short text signals, ensuring that the model retains long text comprehension while learning concise and discriminative information. In the refinement phase, DGCL preserves a small proportion of long text supervision with perturbations, preventing the forgetting of complex semantics and improving robustness. Through the synergy of KPS and DGCL, the framework achieves a smooth transition from comprehensive semantic understanding to concise representation, thereby realizing a dynamic balance between long and short texts and facilitating unified multi-granularity representation learning.}
\par
\textbf{Warm-up Phase.} During the first training stage, the weighting factor $\alpha$ is fixed at 1.0, meaning that optimization relies solely on supervision from long text signals. The objective of this phase is to fully leverage the pretrained advantages of KPS in long text semantic modeling, further consolidating the model ability to represent long text information and laying a solid foundation for the subsequent introduction of short text supervision.
\par
\textbf{Decay Phase.} In the second training stage, $\alpha$ is gradually reduced from 1.0 to a lower bound using a cosine annealing schedule, thereby progressively decreasing the influence of long text supervision. The weight adjustment is computed as follows:
\begin{equation}
\alpha(t)=\alpha_{\min}
           +\frac{1}{2}\,(\alpha_{\text{start}}-\alpha_{\min})
           \Bigl[1+\cos\!\Bigl(\pi\,\dfrac{t-t_{1}}{t_{2}-t_{1}}\Bigr)\Bigr]
\end{equation}
where $t$ denotes the current training step, and $t_1$ and $t_2$ represent the start and end steps of the decay phase, respectively. This phase aims to balance the optimization focus between long and short texts, enhancing the model capability to perceive and represent short text features and facilitating a smooth transition from long text dominance to collaborative optimization.
\par
\textbf{Refinement Phase.} In the final phase, $\alpha$ is maintained at approximately 0.2, and a gradually decaying perturbation term is introduced to improve the model robustness and adaptability to multi-granularity semantics. The perturbation-enhanced weight is computed as follows:
\begin{equation}
\alpha \left( t \right) =\alpha _{\min}+\epsilon \left( t \right) ,\quad \epsilon \left( t \right) \sim U\left( -\delta \left( 1-\hat{t} \right) ,\delta \left( 1-\hat{t} \right) \right) 
\end{equation}
The term $\epsilon$ represents a perturbation sampled from a uniform distribution, while $\delta$ denotes the maximum perturbation amplitude, and $\hat{t}=\frac{t-t_2}{T-t_2}$ refers to the normalized training progress. As training advances, the perturbation gradually decays, effectively preventing the model from entering a saturated state in the later training stages. This mechanism further enhances the model's ability to achieve unified representation and improved generalization across multi-granularity semantic information.
\par
{In summary, by integrating KPS with DGCL, the framework achieves a smooth transition from long text dominant optimization to collaborative optimization of both long and short texts through a structured three-stage training process. KPS extends the text encoder’s capacity and ensures effective modeling of semantic dependencies among long text tokens, while DGCL dynamically adjusts the supervision balance to preserve long text comprehension and enhance short text adaptability. Together, these mechanisms effectively improve the model’s adaptability and robustness in complex semantic understanding and expression tasks.}
\section{Experiments and analysis}
\subsection{Pretraining Setup}
In this study, we pretrained two model sizes of different sizes, ViT-B/16 and ViT-L/14, on the proposed DGTRSD using KPS and DGCL. For text data processing, KPS interpolation ratio was set to 4. As a result, according to Equation(\ref{eq4}), the maximum encodable text token length was extended from 77 to 248. For image data, the standard preprocessing pipeline used in CLIP was employed to maintain consistency. All experiments were conducted on eight NVIDIA RTX 3090 GPUs, and the models were trained for three epochs on DGTRSD under this optimier setting.
\begin{table}[htbp]
\caption{optimier setting}
\centering
\renewcommand{\arraystretch}{1.3}
\sisetup{group-separator = {,}, group-minimum-digits = 4}
\begin{tabular}{>{\centering\arraybackslash}m{2cm} >{\raggedleft\arraybackslash}m{2cm}}
\toprule
\textbf{Parameter} & \textbf{Value} \\
\midrule
optimizer & SGD \\
learning rate & 5.0e-4 \\
momentum & 0.9 \\
dampening & 0.1 \\
\bottomrule
\end{tabular}
\label{opti}
\end{table}

\begin{table*}[htbp]
\centering
\caption{LTCR PERFORMANCE on RSITMD-L and RSICD-L datasets. Results show that Long-CLIP-sft, finetuned on DGTRSD, significantly outperforms Long-CLIP, demonstrating the benefit of domain-adaptive pretraining. DGTRS-CLIP further improves upon Long-CLIP-sft, highlighting the effectiveness of DGCL in enhancing semantic modeling and generalization. The best results are highlighted in \textbf{bold}.}
\label{LTCR}
\renewcommand{\arraystretch}{1.5}
\begin{tabularx}{\textwidth}{>{\centering\arraybackslash}X >{\centering\arraybackslash}m{1.5cm} >{\centering\arraybackslash}X ccc ccc c}
\toprule
\multirow{2}{*}{Testing Dataset} & \multirow{2}{*}{Size} & \multirow{2}{*}{Method} 
& \multicolumn{3}{c}{Text to Image} 
& \multicolumn{3}{c}{Image to Text} 
& \multirow{2}{*}{mR} \\
\cmidrule{4-9}
& & & R@1 & R@5 & R@10 & R@1 & R@5 & R@10 & \\
\midrule

\multirow{6}{*}{RSITMD-L}
& \multirow{3}{*}{ViT-B/16} & Long-CLIP     & 19.03 & 51.33 & 71.02 & 18.14 & 51.33 & 71.02 & 45.43 \\
&                         & Long-CLIP-sft & 38.50 & 73.89 & 84.96 & 36.28 & 70.80 & 84.29 & 64.79 \\
&                         & DGTRS-CLIP       & 39.38 & 70.58 & 86.95 & 37.17 & \textbf{74.12} & \textbf{84.73} & 65.49 \\
\cmidrule{2-10}
& \multirow{3}{*}{ViT-L/14} & Long-CLIP     & 24.12 & 59.29 & 77.88 & 17.70 & 50.44 & 65.93 & 49.23 \\
&                         & Long-CLIP-sft & 39.16 & 72.12 & \textbf{87.39} & 35.62 & 70.58 & 85.40 & 65.04 \\
&                         & DGTRS-CLIP       & \textbf{40.49} & \textbf{74.78} & \textbf{87.39} & \textbf{38.05} & 72.35 & 84.07 & \textbf{66.19} \\
\midrule

\multirow{6}{*}{RSICD-L}
& \multirow{3}{*}{ViT-B/16} & Long-CLIP     & 11.89 & 31.93 & 46.20 & 7.96  & 25.53 & 39.43 & 27.15 \\
&                         & Long-CLIP-sft & 23.15 & 51.51 & 66.61 & 20.22 & 47.67 & 63.68 & 45.47 \\
&                         & DGTRS-CLIP       & 24.06 & 55.44 & 70.54 & \textbf{23.15} & \textbf{52.15} & \textbf{68.62} & 48.99 \\
\cmidrule{2-10}
& \multirow{3}{*}{ViT-L/14} & Long-CLIP     & 14.91 & 36.87 & 52.52 & 6.77  & 24.70 & 38.52 & 29.05 \\
&                         & Long-CLIP-sft & 26.72 & 55.54 & 70.45 & 21.96 & 48.95 & 66.42 & 48.34 \\
&                         & DGTRS-CLIP       & \textbf{27.17} & \textbf{57.46} & \textbf{72.19} & 22.96 & 51.33 & 66.06 & \textbf{49.53} \\
\bottomrule
\end{tabularx}
\end{table*}
\begin{table*}[htbp]
\centering
\caption{STCR PERFORMANCE ON RSITMD AND RSICD DATASETS. Long-CLIP-sft, finetuned on DGTRSD, achieves substantial improvements over Long-CLIP, verifying the effectiveness of domain-adaptive pretraining. Furthermore, DGTRS-CLIP consistently outperforms both Long-CLIP-sft and GeoRSCLIP across most settings, demonstrating the advantage of DGCL in balancing multi-granularity semantic learning and enhancing short text retrieval performance. The best results are highlighted in \textbf{bold}.}
\label{STCR}
\renewcommand{\arraystretch}{1.5}
\begin{tabularx}{\textwidth}{>{\centering\arraybackslash}X >{\centering\arraybackslash}m{1.5cm} >{\centering\arraybackslash}X ccc ccc c}
\toprule
\multirow{2}{*}{Testing Dataset} & \multirow{2}{*}{\textbf{Size}} & \multirow{2}{*}{Method} 
& \multicolumn{3}{c}{Text to Image} 
& \multicolumn{3}{c}{Image to Text} 
& \multirow{2}{*}{mR} \\
\cmidrule(lr){4-6} \cmidrule(lr){7-9}
& & & R@1 & R@5 & R@10 & R@1 & R@5 & R@10 & \\
\midrule

\multirow{9}{*}{RSITMD}
& \multirow{2}{*}{ViT-B/32} & CLIP         & 8.81 & 27.88 & 43.19 & 9.51 & 23.01 & 32.74 & 24.19 \\
&                          & GeoRSCLIP    & 14.16 & 42.39 & 57.52 & 19.03 & 34.51 & 46.46 & 35.68 \\
\cmidrule(lr){2-10}
& \multirow{3}{*}{ViT-B/16} & Long-CLIP    & 9.07 & 33.23 & 49.16 & 8.98 & 24.11 & 33.41 & 26.33 \\
&                          & Long-CLIP-sft & 14.87 & 37.61 & 52.12 & 21.46 & 40.71 & 50.22 & 36.17 \\
&                          & DGTRS-CLIP       & 15.18 & 40.80 & 55.58 & 21.90 & 39.82 & \textbf{52.21} & 37.58 \\
\cmidrule(lr){2-10}
& \multirow{5}{*}{ViT-L/14} & CLIP         & 10.13 & 31.64 & 46.24 & 10.84 & 28.10 & 36.50 & 27.24 \\
&                          & GeoRSCLIP    & 17.35 & 41.46 & 56.24 & 20.35 & 38.05 & 49.56 & 37.17 \\
&                          & Long-CLIP    & 12.74 & 36.99 & 53.27 & 11.28 & 27.43 & 38.50 & 30.04 \\
&                          & Long-CLIP-sft & \textbf{17.52} & 41.68 & 56.73 & 21.02 & 39.38 & \textbf{52.21} & 38.09 \\
&                          & DGTRS-CLIP       & 17.35 & \textbf{43.27} & \textbf{58.10} & \textbf{22.12} & \textbf{42.26} & 51.99 & \textbf{39.18} \\
\midrule

\multirow{9}{*}{RSICD}
& \multirow{2}{*}{ViT-B/32} & CLIP         & 5.78 & 17.73 & 27.76 & 5.31 & 14.18 & 23.70 & 15.74 \\
&                          & GeoRSCLIP    & 9.52 & 27.37 & 40.99 & 11.53 & 28.55 & 39.16 & 26.18 \\
\cmidrule(lr){2-10}
& \multirow{3}{*}{ViT-B/16} & Long-CLIP    & 6.72 & 20.62 & 32.19 & 7.69 & 20.13 & 29.19 & 19.42 \\
&                          & Long-CLIP-sft & 9.11 & 27.47 & 39.71 & 13.08 & 29.83 & 39.62 & 26.47 \\
&                          & DGTRS-CLIP       & 9.92 & 28.60 & 41.87 & 11.53 & 29.19 & 40.99 & 27.01 \\
\cmidrule(lr){2-10}
& \multirow{5}{*}{ViT-L/14} & CLIP         & 4.96 & 18.72 & 29.66 & 6.59 & 17.75 & 28.09 & 17.63 \\
&                          & GeoRSCLIP    & 9.97 & 28.18 & 42.10 & 12.72 & 28.82 & 41.17 & 27.16 \\
&                          & Long-CLIP    & 7.14 & 23.48 & 35.55 & 8.78 & 20.40 & 28.82 & 20.70 \\
&                          & Long-CLIP-sft & 10.01 & 28.60 & 43.73 & \textbf{15.65} & \textbf{31.38} & \textbf{41.26} & 28.44 \\
&                          & DGTRS-CLIP       & \textbf{11.22} & \textbf{31.86} & \textbf{45.49} & 14.36 & 28.36 & 40.53 & \textbf{28.64} \\
\bottomrule
\end{tabularx}
\end{table*}
\begin{table*}[htbp]
\centering
\caption{IC PERFORMANCE ON SIX REMOTE SENSING SCENE CLASSIFICATION DATASETS. Long-CLIP-sft and DGTRS-CLIP significantly improves over general domain models. Despite limited average gains in IC due to coarse-grained semantics, DGTRS-CLIP surpasses GeoRSCLIP in most cases, confirming its advantage in unified semantic modeling and generalization. The best results are highlighted in \textbf{bold}.}
\label{IC}
\renewcommand{\arraystretch}{1.5}
\begin{tabularx}{\textwidth}{>{\centering\arraybackslash}X >{\centering\arraybackslash}X 
>{\centering\arraybackslash}X >{\centering\arraybackslash}X >{\centering\arraybackslash}X 
>{\centering\arraybackslash}X >{\centering\arraybackslash}X >{\centering\arraybackslash}X c}
\toprule
\multirow{2}{*}{Size} & \multirow{2}{*}{Method} 
& \multicolumn{6}{c}{Testing Dataset} 
& \multirow{2}{*}{Mean} \\
\cmidrule(lr){3-8}
& & RESISC45 & EuroSAT & PatternNet & WHU-RS19 & OPTIMAL-31 & RS\_C11 & \\
\midrule

\multirow{3}{*}{ViT-B/32} & CLIP         & 53.23 & 47.92 & 49.83 & 73.13 & 67.10 & 30.60 & 53.64 \\
                          & RemoteCLIP   & 70.08 & 30.42 & 59.53 & 95.42 & 79.57 & 54.87 & 64.98 \\
                          & GeoRSCLIP    & 72.82 & 58.99 & 76.95 & 72.82 & 82.69 & 57.63 & 70.32 \\
\cmidrule(lr){1-9}
\multirow{3}{*}{ViT-B/16} & Long-CLIP    & 61.83 & 51.57 & 58.68 & 80.50 & 72.04 & 58.20 & 63.80 \\
                          & Long-CLIP-sft & 73.12 & 57.55 & 74.30 & 87.16 & 85.27 & 76.79 & 75.70 \\
                          & DGTRS-CLIP    & 73.62 & 48.51 & 74.63 & 88.36 & 83.60 & \textbf{71.51} & 73.37 \\
\cmidrule(lr){1-9}
\multirow{6}{*}{ViT-L/14} & CLIP         & 61.57 & 24.73 & 61.32 & 82.29 & 67.10 & 47.48 & 57.41 \\
                          & RemoteCLIP   & 74.97 & 39.07 & 63.48 & \textbf{92.04} & 81.99 & 68.18 & 69.95 \\
                          & GeoRSCLIP    & 73.16 & 61.77 & 74.23 & 88.66 & 82.80 & 67.45 & 74.68 \\
                          & Long-CLIP    & 65.79 & 61.33 & 70.37 & 87.16 & 76.34 & 57.87 & 69.81 \\
                          & Long-CLIP-sft & 75.30 & \textbf{58.17} & 76.61 & 88.26 & 86.18 & 69.97 & \textbf{75.75} \\
                          & DGTRS-CLIP    & \textbf{76.32} & 55.29 & \textbf{76.72} & 89.95 & \textbf{87.96} & 69.97 & 75.65 \\
\bottomrule
\end{tabularx}
\end{table*}
\subsection{Downstream Task Benchmark}
We conducted four cross-modal tasks to evaluate the transfer and generalization capabilities of DGTRS-CLIP, namely: Long text Cross-modal Retrieval (LTCR), Short text Cross-modal Retrieval (STCR), Image Classification (IC) and Semantic Localization (SeLo). Among them, LTCR and STCR are used to assess the cross-modal alignment of long texts and short texts with remote sensing images, IC evaluates the zero-shot transfer ability of DGTRS-CLIP, and SeLo evaluates DGTRS-CLIP fine-grained understanding and alignment of local information. 
\par
\textbf{1) LTCR: }In the domain of remote sensing, there is currently no benchmark for cross-modal retrieval of long texts. Based on the short-text cross-modal retrieval datasets RSICD and RSITMD, we constructed the first such benchmark, namely RSITMD-L and RSICD-L, using Qwen2-VL with a Short-to-Long Text Generation approach. The final RSITMD-L and RSICD-L datasets pair images with five short texts from the original datasets and one long text generated by Qwen2-VL. Using DGTRS-CLIP as the backbone model, we conducted retrieval tests from long text to image and from image to long text on the RSICD and RSITMD test sets, as partitioned by Liu et al~\cite{liu2024remoteclip}. Evaluation metrics include recall\@1/5/10 and mean recall.

\textbf{2) STCR: }Similar to LTCR, we used the same setup for short text-to-image and image-to-short text retrieval tasks on the RSICD and RSITMD test sets, with the same evaluation metrics of recall@1/5/10 and mean recall.

\textbf{3) IC: }CLIP's zero-shot image classification is achieved by embedding class names into prompt templates, such as "a photo of a \{class\_name\}". By matching images with prompts for each class, the one with the highest similarity is selected as the output category. Following CLIP's prompt template, we use "a satellite photo of \{class\_name\}" for our prompts and test on six scene classification datasets: RESISC45~\cite{cheng2017remote}, EuroSAT~\cite{helber2019eurosat}, PatternNet\cite{zhou2018patternnet}, WHU-RS19~\cite{zhao2015dirichlet}, OPTIMAL-31~\cite{wang2018scene}, and RS\_C11~\cite{zhao2016feature} (including both training and testing sets). The evaluation metric is accuracy.

\textbf{4) SeLo: }Building on the work of Yuan et al.~\cite{yuan2022learning}, we replaced DGTRS-CLIP encoder with a backbone and conducted SeLo task testing on AIR-SLT. The evaluation metrics include: \( R\textsubscript{su} \): Measures the probability ratio between the ground-truth (GT) region and other regions. When the model focuses attention on the GT region, the value is close to 1; otherwise, it is close to 0. \( R\textsubscript{as} \): Quantifies the distance between the GT region center and the closest attention area. The value is close to 0 when attention is close to the GT center, and close to 1 when it is not. \( R\textsubscript{da} \): Measures the divergence of the model attention, which assesses retrieval model stability and whether the image contains the query target or relationship. \( R\textsubscript{mi} \): A composite measure incorporating the three above metrics: 
\begin{equation}
R_{mi}=\omega _{su}R_{su}+\omega _{as}\left( 1-R_{as} \right) +\omega _{da}R_{da}
\end{equation}
where $\omega_{su} = 0.4$, $\omega_{as} = 0.35$, $\omega_{da} = 0.25$. Except for $R_{as}$, all other metrics are better when higher.
\par
To evaluate the performance of DGTRS-CLIP in these downstream tasks, we compared it with CLIP[10], Long-CLIP[23], and remote sensing-based VLFM models GeoRSCLIP~\cite{zhang2024rs5m} and RemoteCLIP~\cite{liu2024remoteclip}. In the experimental results, CLIP and Long-CLIP represent VLFM models trained in the general domain, while GeoRSCLIP and RemoteCLIP are VLFM models trained on remote sensing short text tasks. Long-CLIP-sft is obtained by directly performing full finetuning on Long-CLIP.
\subsection{Main Experimental Results}
TABLE.~\ref{LTCR} reports the LTCR results on RSITMD-L and RSICD-L using three models, Long-CLIP, Long-CLIP-sft, and DGTRS-CLIP, with ViT-B/16 and ViT-L/14 as backbones. Compared to the original Long-CLIP, Long-CLIP-sft achieves substantial performance improvements across all metrics. Notably, R@1 and mR show average gains of approximately 15 to 20 percentage points, highlighting the effectiveness of domain-adaptive data in improving cross-modal retrieval. Building upon this, DGTRS-CLIP further enhances retrieval performance through the introduction of DGCL strategy. Consistent improvements over Long-CLIP-sft are observed across both datasets, demonstrating that DGCL contributes to more effective multi-granularity semantic modeling and strengthens the model's generalization capability.
\par
\begin{figure}[htbp]
  \centering
  \includegraphics[width=\linewidth]{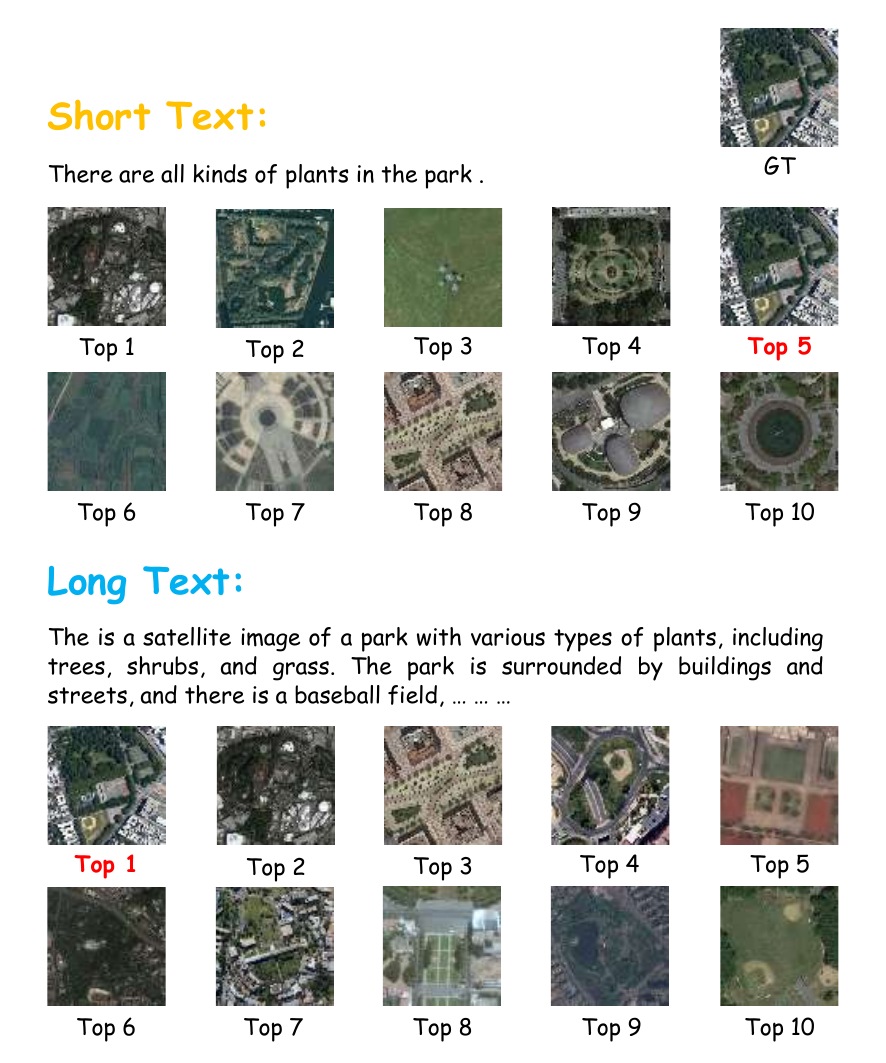}
  \caption{Top-10 retrieval results using the short text and long text of GT, where the correct result is highlighted in bold red. As shown, although the retrieved images exhibit certain semantic similarity under both settings, the correct image appears at top-1 with long text but only at top-5 with short text, indicating the inherent ambiguity of short text and its impact on retrieval accuracy.}
  \label{retrieval}
\end{figure}

TABLE.~\ref{STCR} reports the STCR results on RSITMD and RSICD datasets across different models and backbones. In addition, Long-CLIP-sft further outperforms Long-CLIP in all metrics, especially in R@1 and mR, confirming the effectiveness of domain-adaptive finetuning for short text retrieval. Building upon this, DGTRS-CLIP achieves consistent performance gains over Long-CLIP-sft across model size. Specifically, DGTRS-CLIP surpasses GeoRSCLIP in most cases, indicating that the Dual-Granularity Curriculum Learning (DGCL) strategy not only improves the modeling ability of long text scenarios, but also improves the model performance on short text retrieval tasks by balancing multi-granularity semantic supervision. {It is also worth noting that the performance improvement on STCR is considerably smaller than that on LTCR. We attribute this to the large-scale characteristics and high scene similarity of remote sensing imagery, where the ground-truth image of a given caption often has multiple visually similar candidates in the retrieval database, thereby increasing the likelihood of incorrect retrievals, as illustrated in Fig.~\ref{retrieval}. The fundamental reason lies in the inherent semantic ambiguity and limited expressiveness of short text. Therefore, aligning remote sensing images with long text is of particular importance, as it can effectively alleviate semantic ambiguity and improve retrieval accuracy.}
\par
\begin{table*}[htbp]
\centering
\caption{SeLo Performance on AIR-SLT dataset. The results include four metrics: $R_{\mathrm{su}}$, $R_{\mathrm{as}}$, $R_{\mathrm{da}}$, and $R_{\mathrm{mi}}$. Higher values indicate better performance except for $R_{\mathrm{as}}$, where lower values are preferred. The results of Long-CLIP-sft and DGCL-CLIP show significant improvements over GeoRSCLIP and RemoteCLIP, indicating that dual-granularity text training greatly aids in local alignment. The best results are highlighted in \textbf{bold}.}
\label{SeLo}
\renewcommand{\arraystretch}{1.5}
\begin{tabularx}{\textwidth}{>{\centering\arraybackslash}m{1.5cm} >{\centering\arraybackslash}m{3cm} 
>{\centering\arraybackslash}X >{\centering\arraybackslash}X >{\centering\arraybackslash}X >{\centering\arraybackslash}X}
\toprule
\multirow{2}{*}{Size} & \multirow{2}{*}{Method} & \multicolumn{4}{c}{Semantic Localization (AIR-SLT)} \\
\cmidrule(lr){3-6}
& & $R_{\mathrm{su}}$$\uparrow$ & $R_{\mathrm{as}}$$\downarrow$ & $R_{\mathrm{da}}$$\uparrow$ & $R_{\mathrm{mi}}$$\uparrow$ \\
\midrule
--- & SeLo-v1      & 0.6920 & 0.3323 & 0.6667 & 0.6772 \\
--- & SeLo-v2      & 0.7199 & 0.2925 & 0.6658 & 0.7021 \\
\cmidrule(lr){1-6}
\multirow{3}{*}{ViT-B/32} 
& CLIP         & 0.7275 & 0.2878 & 0.7062 & 0.7168 \\
& RemoteCLIP   & 0.7365 & 0.3008 & 0.6928 & 0.7125 \\
& GeoRSCLIP    & 0.7546 & 0.2610 & 0.7180 & 0.7400 \\
\cmidrule(lr){1-6}
\multirow{3}{*}{ViT-B/16} 
& Long-CLIP    & 0.7508 & 0.2405 & 0.7369 & 0.7503 \\
& Long-CLIP-sft & 0.7689 & \textbf{0.2371} & 0.7345 & 0.7582 \\
& \textbf{DGTRS-CLIP}      & 0.7795 & 0.2440 & 0.7668 & \textbf{0.7681} \\
\cmidrule(lr){1-6}
\multirow{6}{*}{ViT-L/14} 
& CLIP         & 0.7318 & 0.3042 & 0.6514 & 0.6991 \\
& RemoteCLIP   & 0.7727 & 0.2917 & 0.6834 & 0.7279 \\
& GeoRSCLIP    & 0.7541 & 0.2513 & 0.7573 & 0.7530 \\
& Long-CLIP    & 0.7638 & 0.2691 & 0.7297 & 0.7438 \\
& Long-CLIP-sft & 0.7803 & 0.2495 & 0.7620 & 0.7653 \\
& \textbf{DGTRS-CLIP}      & \textbf{0.7827} & 0.2494 & \textbf{0.7676} & 0.7677 \\
\bottomrule
\end{tabularx}
\end{table*}
TABLE.~\ref{IC} summarizes IC results on six remote sensing scene classification datasets. Overall, both Long-CLIP-sft and DGTRS-CLIP and DGTRS-CLIP significantly outperform the general domain models, CLIP and Long-CLIP. Specifically, Long-CLIP-sft and DGTRS-CLIP achieves substantial accuracy improvements across all datasets, demonstrating the effectiveness of domain-specific pretraining for remote sensing classification tasks. Particularly on multi-class scenes such as RESISC45 and OPTIMAL-31. However, compared with the LTCR and STCR tasks, the average accuracy gains of DGTRS-CLIP in IC are relatively limited. This can be attributed to the nature of IC, which primarily involves scene-level classification with coarse-grained semantics. In such scenarios, short text representations are often sufficient to capture key discriminative information. The incorporation of long text supervision, while enhancing semantic richness, may introduce redundant information in simple datasets (e.g., EuroSAT), which could slightly affect decision-making effectiveness. However, DGTRS-CLIP still achieves better or comparable performance to GeoRSCLIP and RemoteCLIP, which are specifically optimized for short text in the remote sensing domain. This further confirms that the proposed multi-granularity semantic optimization strategy improves not only long text modeling, but also enhances unified semantic representation and cross-dataset generalization in image classification tasks.
\par
TABLE~\ref{SeLo} presents the experimental results of SeLo on AIR-SLT, compared to the baseline models SeLo-v1[33] and SeLo-v2[84]. In both ViT-B/16 and ViT-L/14 configurations, DGTRS-CLIP achieves a significant improvement over other models, particularly in the Rmi metric, demonstrating the advantage of DGTRS-CLIP in remote sensing image localization tasks. This improvement highlights that DGCL strengthens multi-granularity semantic modeling, making DGTRS-CLIP more robust and accurate in target area focus and fine-grained alignment.
Notably, while Long-CLIP was not trained on remote sensing data, it still outperforms the original CLIP model in most metrics due to its ability to process long texts. By incorporating DGCL, DGTRS-CLIP further enhances performance, especially in target localization and image-text alignment tasks. Compared to GeoRSCLIP and RemoteCLIP, which are trained with short text data, DGTRS-CLIP consistently outperforms them across all metrics, demonstrating that KPS and DGCL help DGTRS-CLIP achieve superior performance in target attention, suppression of irrelevant interference, and fine-grained image-text alignment.
\par
\begin{figure}[htbp]
  \centering
  \includegraphics[width=\linewidth]{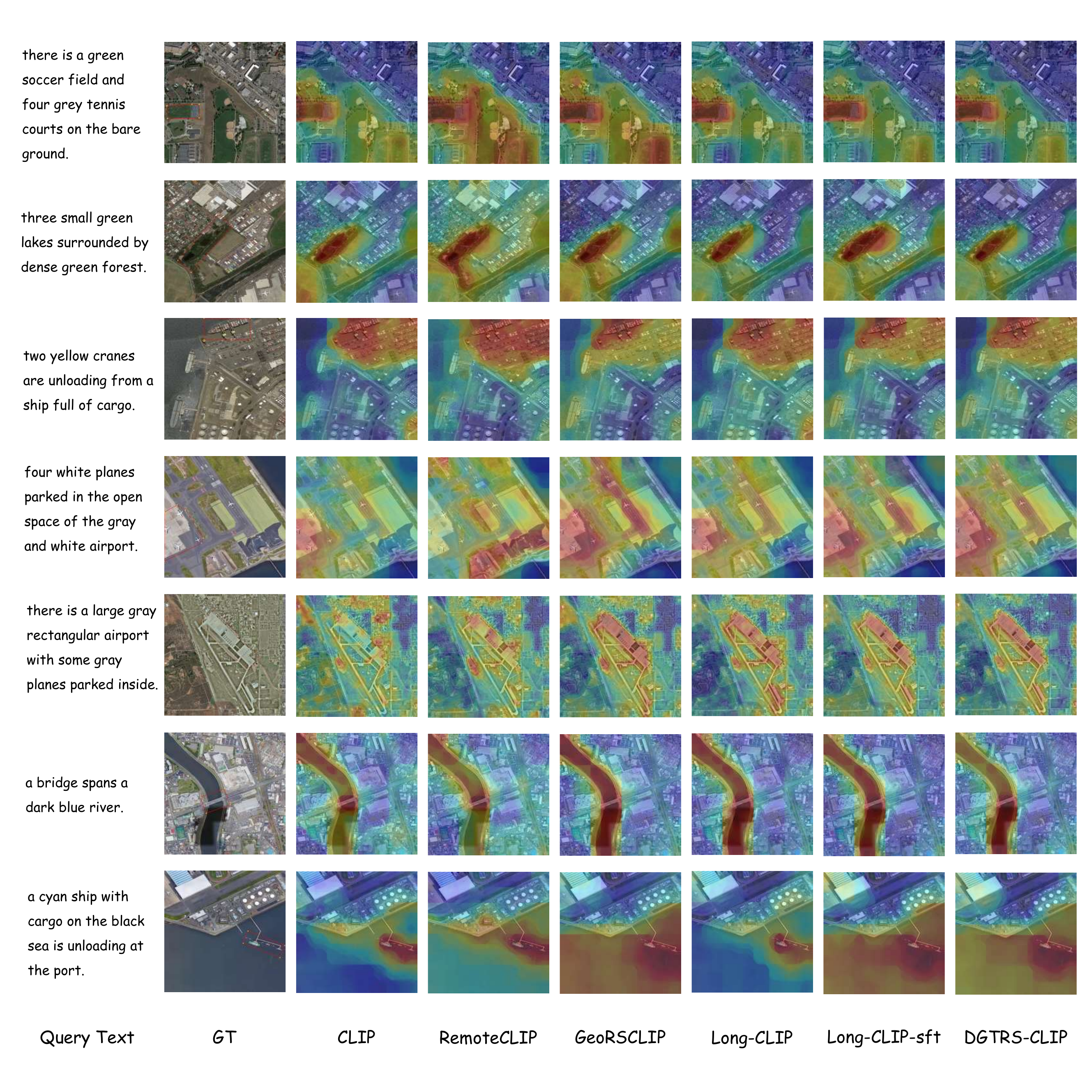}
  \caption{Aggregated heatmap results of different methods on the SeLo task. DGTRS-CLIP achieves more precise attention coverage of the GT regions, and models trained with both long and short texts (e.g., DGTRS-CLIP, Long-CLIP) reduce false activations compared to short-text-only models. However, DGTRS-CLIP may still exhibit semantic confusion in complex queries, as illustrated in the last two rows.}
  \label{selo}
\end{figure}

{Fig.~\ref{selo} presents the aggregated heatmap results of different methods on the SeLo task. It can be observed that, compared with other approaches, DGTRS-CLIP produces attention maps that not only cover the GT region but also exhibit finer granularity in some cases. Moreover, models trained with both long and short texts (e.g., Long-CLIP, Long-CLIP-sft, DGTRS-CLIP) demonstrate fewer false activations on irrelevant regions than those trained solely with short texts (e.g., CLIP, RemoteCLIP, GeoRSCLIP). For instance, when using the query “there is a green soccer field and four grey tennis courts on the bare ground”, models trained with both granularities concentrate their attention exclusively within the GT region, whereas short-text-only models also attend to the playground in the bottom-right corner. This indicates that incorporating both granularities provides richer semantic alignment while retaining the ability of short text to capture coarse-grained information. Nevertheless, DGTRS-CLIP still exhibits certain limitations. As shown in the last two rows of Fig.~\ref{selo}, DGTRS-CLIP sometimes simultaneously attends to both the subject and object in the query text, reflecting a degree of semantic confusion. We argue that this issue is not caused by the use of long texts, since Long-CLIP still produces accurate results in these cases. Interestingly, RemoteCLIP performs better under such scenarios, which we attribute to the specific linguistic style of its training corpus. Because RemoteCLIP captions are synthesized via M2B and B2C template-based generation, the semantic structures are more regular, making them easier for the model to parse. However, such fixed language styles are unlikely to match real-world scenarios. Therefore, introducing more diverse training corpora and incorporating longer token modeling will be an effective direction to address this limitation.}
\par
{Among the four downstream tasks, LTCR and STCR evaluate the global semantic alignment between remote sensing images and texts at different granularities, with LTCR focusing on long text supervision and STCR emphasizing short text supervision. The IC task assesses coarse-grained image–text matching from the perspective of overall image classification, while SeLo examines semantic localization at the patch level. Overall, compared with existing open-source methods, DGTRS-CLIP achieves state-of-the-art performance across these tasks, demonstrating its dual-granularity semantic extraction capability and strong adaptability to diverse downstream applications.}
\subsection{Ablation Study}
\subsubsection{Effectiveness Evaluation of KPS and DGCL}
From the experimental results presented in TABLE~\ref{tab:kps_dgcl_ablation}, it can be observed that both KPS and DGCL significantly improve the model's performance in cross-modal retrieval and semantic understanding tasks. KPS enhances the alignment between long text and images, thereby boosting the model cross-modal retrieval capability. On the other hand, DGCL guides the model in handling fine-grained text information, improving its performance in multi-granularity retrieval and SeLo. The optimal performance is achieved when KPS and DGCL are used in combination, indicating that the integration of both components maximizes the overall model performance. Therefore, KPS and DGCL are critical factors in enhancing model performance, and their combined use provides effective support for cross-modal understanding and retrieval tasks.
\begin{table}[htbp]
\caption{Impact of KPS and DGCL on Performance}
\centering
\renewcommand{\arraystretch}{1.2}
\setlength{\tabcolsep}{4pt}
\begin{tabularx}{\linewidth}{cc *{6}{>{\centering\arraybackslash}X}}
\toprule
\textbf{KPS} & \textbf{DGCL} 
& \makecell{\textbf{LT2I}\\\textbf{R@1}} 
& \makecell{\textbf{I2LT}\\\textbf{R@1}} 
& \makecell{\textbf{ST2I}\\\textbf{R@1}} 
& \makecell{\textbf{I2RT}\\\textbf{R@1}} 
& \makecell{\textbf{Mean}\\\textbf{Acc.}} 
& \textbf{$\boldsymbol{R}_{\mathrm{mi}}$} \\
\midrule
\ding{55} & \ding{55} & 30.62 & 31.64 & 14.87 & 19.91 & 74.80 & 0.7424 \\
\checkmark & \ding{55} & 35.40 & 36.28 & 15.13 & 20.12 & 76.99 & 0.7427 \\
\ding{55} & \checkmark & 33.41 & 33.63 & 13.36 & 17.04 & 71.22 & 0.7571 \\
\checkmark & \checkmark & 39.38 & 37.17 & 15.18 & 21.90 & 73.37 & 0.7681 \\
\bottomrule
\end{tabularx}
\label{tab:kps_dgcl_ablation}
\end{table}
\subsubsection{Impact of Dataset Evaluation}
From the experimental results presented in TABLE~\ref{tab:evaluation_ablation}, it can be observed that the introduction of the evaluation phase did not lead to significant changes in the model performance. Although there were slight improvements in certain tasks, such as ST2I and I2RT, the evaluation did not result in substantial improvements for the cross-modal retrieval tasks (LT2I and I2LT). The Mean Accuracy decreased, indicating that the evaluation phase had a limited impact on the overall performance. Despite the lack of significant performance improvement after introducing the evaluation, we believe that this result is likely due to the fact that only about 7\% of the data was discarded during the filtering process. This suggests that most of the data was not excluded and therefore the impact of the evaluation phase was relatively small. A more detailed discussion on the quality of DGTRSD will be provided in \textbf{Discussion}.

\begin{table}[htbp]
\caption{Impact of Evaluation on Performance}
\centering
\renewcommand{\arraystretch}{1.2}
\setlength{\tabcolsep}{3pt}
\begin{tabularx}{\linewidth}{c *{6}{>{\centering\arraybackslash}X}}
\toprule
\textbf{Evaluation} 
& \makecell{\textbf{LT2I}\\\textbf{R@1}} 
& \makecell{\textbf{I2LT}\\\textbf{R@1}} 
& \makecell{\textbf{ST2I}\\\textbf{R@1}} 
& \makecell{\textbf{I2RT}\\\textbf{R@1}} 
& \makecell{\textbf{Mean}\\\textbf{Acc.}} 
& \textbf{$\boldsymbol{R}_{\mathrm{mi}}$} \\
\midrule
\ding{55}      & 39.82 & 36.73 & 13.98 & 21.24 & 75.21 & 0.7596 \\
\checkmark     & 39.38 & 37.17 & 15.18 & 21.90 & 73.37 & 0.7681 \\
\bottomrule
\end{tabularx}
\label{tab:evaluation_ablation}
\end{table}
\subsubsection{Impact of Learning Order in DGCL}
By comparing the "From Short to Long" and "From Long to Short" learning orders in TABLE~\ref{tab:dgcl_cross_grain}, it can be observed that the "From Long to Short" learning order leads to a significant improvement in the overall performance of the model, particularly in cross-modal retrieval tasks. This order notably enhances the alignment between long text and images and also brings certain improvements in the short text-to-image retrieval task. This suggests that the "Long to Short" learning order is more favorable for cross-modal understanding, particularly when addressing the alignment between long text and images, as it better captures semantic information and facilitates effective alignment.
\begin{table}[htbp]
\caption{Impact of DGCL Learning Order on Alignment Performance}
\centering
\renewcommand{\arraystretch}{1.2}
\setlength{\tabcolsep}{3pt}
\begin{tabularx}{\linewidth}{c *{6}{>{\centering\arraybackslash}X}}
\toprule
\textbf{Learning Order} 
& \makecell{\textbf{LT2I}\\\textbf{R@1}} 
& \makecell{\textbf{I2LT}\\\textbf{R@1}} 
& \makecell{\textbf{ST2I}\\\textbf{R@1}} 
& \makecell{\textbf{I2RT}\\\textbf{R@1}} 
& \makecell{\textbf{Mean}\\\textbf{Acc.}} 
& \textbf{$\boldsymbol{R}_{\mathrm{mi}}$} \\
\midrule
From Short to Long & 35.40 & 33.41 & 14.51 & 18.36 & 75.20 & 0.7442 \\
From Long to Short & 39.38 & 37.17 & 15.18 & 21.90 & 73.37 & 0.7681 \\
\bottomrule
\end{tabularx}
\label{tab:dgcl_cross_grain}
\end{table}

\subsubsection{Impact of Text Granularity on Pretraining Performance}
To evaluate the effectiveness of our multi-granularity text pretraining, we compared the results of different granularity text learning. {TABLE~\ref{tab:granularity_time} presents the time required for both pretraining and evaluation. The results show that using dual-granularity text does not significantly affect the computational speed of the model.} As shown in TABLE~\ref{tab:granularity_ablation}, dual-granularity pretraining using both short and long texts outperforms single-granularity pretraining using only short or long texts across all tasks. Specifically, models pretrained with long texts exhibit superior R@1 results in LTCR and STCR, demonstrating that longer texts indeed contribute to better alignment between image and text. This also indicates that KPS can effectively extend the model’s text encoding capacity without affecting the alignment between short texts and images. However, in the IC and SeLo tasks, models pretrained with short texts perform better than those pretrained with long texts. We hypothesize that this is because, during pretraining, the model may lose the ability to match key information in short texts and instead focus more on global modeling.
\begin{table}[htbp]
\caption{Pretraining and evaluation time under different granularity settings. Pretraining is tested on ViT-B/16; LTCR is tested on RSITMD-L; STCR is tested on RSITMD; IC is tested on all test sets used in the paper; and SeLo is tested on AIR-SLT.}
\label{tab:granularity_time}
\centering
\renewcommand{\arraystretch}{1.1}
\setlength{\tabcolsep}{3pt}
\begin{tabular}{m{2.1cm}<{\centering} m{2.2cm}<{\centering} cccc}
\toprule
\multirow{2}{*}{\textbf{Granularity}} & 
\multirow{2}{*}{\textbf{Pretrain}} & 
\multicolumn{4}{c}{\textbf{Evaluation}} \\
\cmidrule(lr){3-6}
 &  & \textbf{LTCR} & \textbf{STCR} & \textbf{IC} & \textbf{SeLo} \\
\midrule
short           & 28.50~min & 5~s & 5~s & 98~s  & 20.50~min \\
long            & 31.08~min & 7~s & 7~s & 104~s & 21.33~min \\
short and long      & 33.50~min & 7~s & 7~s & 104~s & 21.30~min \\
\bottomrule
\end{tabular}
\end{table}

\begin{table}[htbp]
\caption{Impact of Text Granularity on Pretraining Performance}
\centering
\renewcommand{\arraystretch}{1.2}
\setlength{\tabcolsep}{3pt}
\begin{tabularx}{\linewidth}{c *{6}{>{\centering\arraybackslash}X}}
\toprule
\textbf{Granularity} 
& \makecell{\textbf{LT2I}\\\textbf{R@1}} 
& \makecell{\textbf{I2LT}\\\textbf{R@1}} 
& \makecell{\textbf{ST2I}\\\textbf{R@1}} 
& \makecell{\textbf{I2RT}\\\textbf{R@1}} 
& \makecell{\textbf{Mean}\\\textbf{Acc.}} 
& \textbf{$\boldsymbol{R}_{\mathrm{mi}}$} \\
\midrule
short           & 25.66 & 26.55 & 12.92 & 17.70 & 72.36 & 0.7402 \\
long            & 35.62 & 34.29 & 15.13 & 19.47 & 71.52 & 0.7152 \\
short and long  & 39.38 & 37.17 & 15.18 & 21.90 & 73.37 & 0.7681 \\
\bottomrule
\end{tabularx}
\label{tab:granularity_ablation}
\end{table}

\section{Discussion}
\subsection{Model Architecture and Task Adaptation}
Despite DGTRS-CLIP excellent performance across several tasks, it currently focuses primarily on cross-modal retrieval and classification tasks, without extending to more fine-grained remote sensing tasks such as object detection, semantic segmentation, and change detection. These tasks typically rely on pixel-level annotations and the modeling of fine spatial relationships, thus requiring more advanced model architectures and higher training demands. In object detection and semantic segmentation, DGTRS-CLIP can be enhanced to better adapt to complex scenes in remote sensing images by improving its localization and segmentation accuracy. In change detection, training the model using multi-temporal data can improve the recognition of land cover changes, thus enhancing the model ability to characterize spatiotemporal dynamics. However, since this study primarily focuses on exploring the alignment issue between long and short text descriptions and remote sensing image, it has not addressed model adaptation for these more fine-grained tasks. While these tasks are critical in remote sensing applications, their implementation typically requires additional data annotations and structural adjustments. Future work could leverage the findings of this study to further optimize DGTRS-CLIP, improving its performance in fine-grained tasks such as object detection, semantic segmentation, and change detection.

\subsection{Dataset Construction}
Regarding dataset construction, although DGTRSD provides strong support for remote sensing image-text understanding tasks, there remains room for further improvement. The dataset should cover a broader range of remote sensing scenes, include images with varying spatial resolutions, and provide more detailed captions. In our IC experiments, analysis of the confusion matrices from the PatternNet and EuroSAT datasets revealed that certain classes with lower recall (e.g., “Herbaceous Vegetation” and “Ferry Terminal”) appeared less frequently in the long text captions. This suggests that these categories were not effectively aligned with textual labels during the pretraining phase, thereby negatively impacting the model performance on these classes.

\begin{table}[htbp]
\caption{Class Word occurrence counts extracted from DGTRSD.}
\centering
\renewcommand{\arraystretch}{1.15}
\sisetup{group-separator = {,}, group-minimum-digits = 4}
\begin{tabular}{>{\raggedright\arraybackslash}p{2.7cm} >{\raggedleft\arraybackslash}p{1.8cm}}
\toprule
\textbf{Class Word} & \textbf{Count} \\
\midrule
road & \num{2240238} \\
closed road & \num{1} \\
\midrule
residential & \num{1079380} \\
sparse residential & \num{129} \\
\midrule
parking & \num{608776} \\
parking space & \num{10155} \\
parking lot & \num{50631} \\
\midrule
terminal & \num{17970} \\
ferry terminal & \num{42} \\
\midrule
crop & \num{242938} \\
annual crop & \num{7625} \\
permanent crop & \num{7060} \\
\midrule
forest & \num{678125} \\
runway & \num{101061} \\
herbaceous vegetation & \num{3} \\
oil well & \num{3801} \\
oil gas field & \num{0} \\
\bottomrule
\end{tabular}
\label{tab:keyword_counts}
\end{table}

Additionally, categories such as “parking lot” and “parking space” were often indistinguishable, indicating a lack of fine-grained attribute information in the training data. As a result, the model tends to recognize only the general concept of “parking,” without the ability to differentiate between more specific subcategories. Therefore, future efforts in dataset construction should aim to incorporate a wider variety of remote sensing scenarios along with detailed attribute-level annotations, in order to enhance the model capability to distinguish among closely related classes.
\begin{figure}[htbp]
  \centering
  \includegraphics[width=\linewidth]{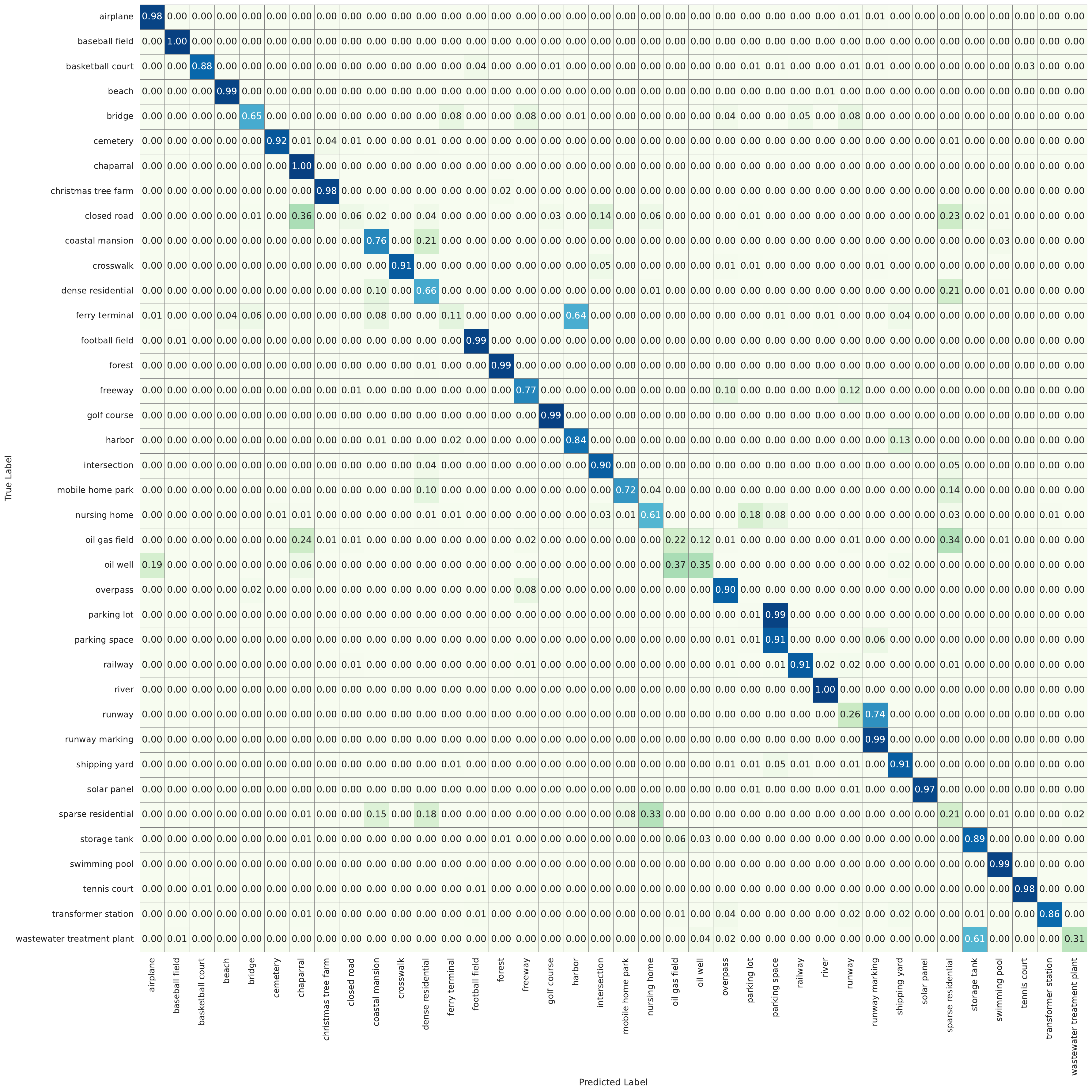}
  \caption{Confusion Matrix for Classification Results on PatternNet.}
  \label{cmpattern}
\end{figure}

\begin{figure}[htbp]
  \centering
  \includegraphics[width=\linewidth]{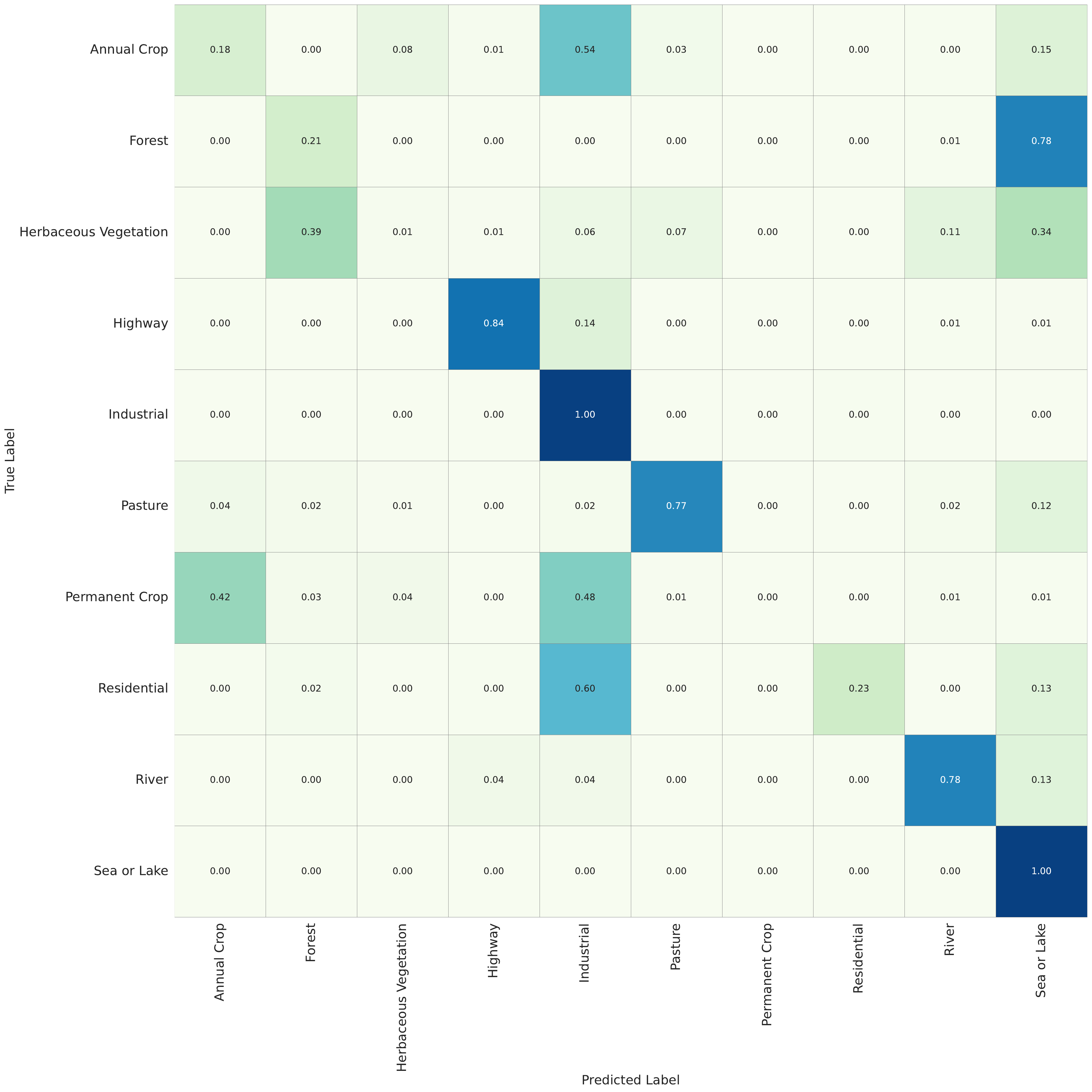}
  \caption{Confusion Matrix for Classification Results on EuroSAT.}
  \label{cmpattern}
\end{figure}
The diversity in image resolution is another aspect that warrants further improvement. Although keywords such as “forest” and “residential” appear with high frequency in DGTRSD, the recall for these classes on EuroSAT remains relatively low. This performance gap may be attributed to the reliance on high-resolution datasets (e.g., LoveDA, DIOR) for caption annotation in this study. Since EuroSAT consists of lower-resolution image, the model struggles to extract meaningful visual features, leading to suboptimal performance even on common categories. This highlights the need to incorporate a broader range of low-resolution training data to enhance the model generalization ability across diverse remote sensing scenarios.

As for the quality of captions, although the average length of long text captions in DGTRSD is 97.03—substantially higher than the 32.06 tokens for short text captions—it still falls short of the model maximum input capacity of 248 tokens. This limitation is primarily due to the relatively short captions in VersaD, which was used for finetuning the Qwen2.5-VL-7B-Instruction. Although the long text format offers more detail overall, it does not fully exploit the model text capacity. Moreover, when using smaller MLLMs without domain-specific adaptation, hallucination issues may arise, with generated captions deviating from actual image content. This is exacerbated by the long-tail nature of remote sensing data, which general domain MLLMs often fail to represent adequately. These models tend to struggle with extracting fine-grained semantic and spatial information from remote sensing images. Future work should therefore prioritize domain adaptation techniques to mitigate hallucination and enhance image-text alignment in this domain.

\subsection{Dataset Evaluation}
We believe that the limitations in the dataset construction process significantly contribute to the scarcity of high-quality annotations, as reflected in the low number of samples rated 5 (only 83 in total). The difficulty of general domain models in generating accurate and detailed remote sensing captions, especially without domain-specific finetuning, limits the production of samples that meet stringent consistency standards. This highlights the need for more refined annotation strategies and domain-adapted models to generate higher-quality image-text pairs, particularly in terms of detailed and spatially coherent long text descriptions.

The results in the TABLE.~\ref{tab2} also show that only the BigEarthNet subset of DGTRSD has a retention rate below 90\% after filtering. We believe this is due to the relatively low resolution of remote sensing images in BigEarthNet. Additionally, we contend that image resolution is one of the factors affecting the quality of remote sensing image datasets when captioning and evaluating using MLLMs. However, we argue that the fundamental cause of this phenomenon lies in the lag in understanding of remote sensing data by general domain MLLMs.

In conclusion, while DGTRS-CLIP and DGTRSD lay a solid foundation for remote sensing image-text understanding, several key areas require further refinement. These include increasing dataset diversity, improving adaptation to fine-grained tasks, enhancing domain-specific caption generation, and addressing limitations related to image resolution and annotation quality. Future efforts in these directions will be crucial to advancing cross-modal learning in remote sensing applications.
\section{Conclusion}
VLFM has demonstrated significant potential in remote sensing image understanding. However, existing methods are primarily trained on short text data, leading to issues such as inadequate modeling of target space relationships and suboptimal attention distribution in the semantic understanding process. To address these challenges, we propose DGTRSD, which includes both long and short text captions, enabling the model to align semantic information at two different granularity. We then introduce DGTRS-CLIP, which comprises two key components: 1) the use of KPS to extend the text encoding length of CLIP, thereby enhancing its ability to model long text; 2) DGCL which dynamically adjusts the alignment of long and short texts with remote sensing images, enabling DGTRS-CLIP to maintain strong long text understanding while preserving short text comprehension. Based on these innovations, we performed a full finetuning of CLIP and achieved state-of-the-art performance across tasks such as cross-modal retrieval, scene classification, and semantic localization. Additionally, we design comprehensive ablation studies to validate the effectiveness of our approach.

\section*{Acknowledgments}
This research was funded by the National Key R\&D Program of China under grant number 2021YFB3900504.
\bibliographystyle{IEEEtran}
\bibliography{ref} 

\clearpage
\vspace{11pt}

\begin{IEEEbiography}[{\includegraphics[width=1in,height=1.25in,clip,keepaspectratio]{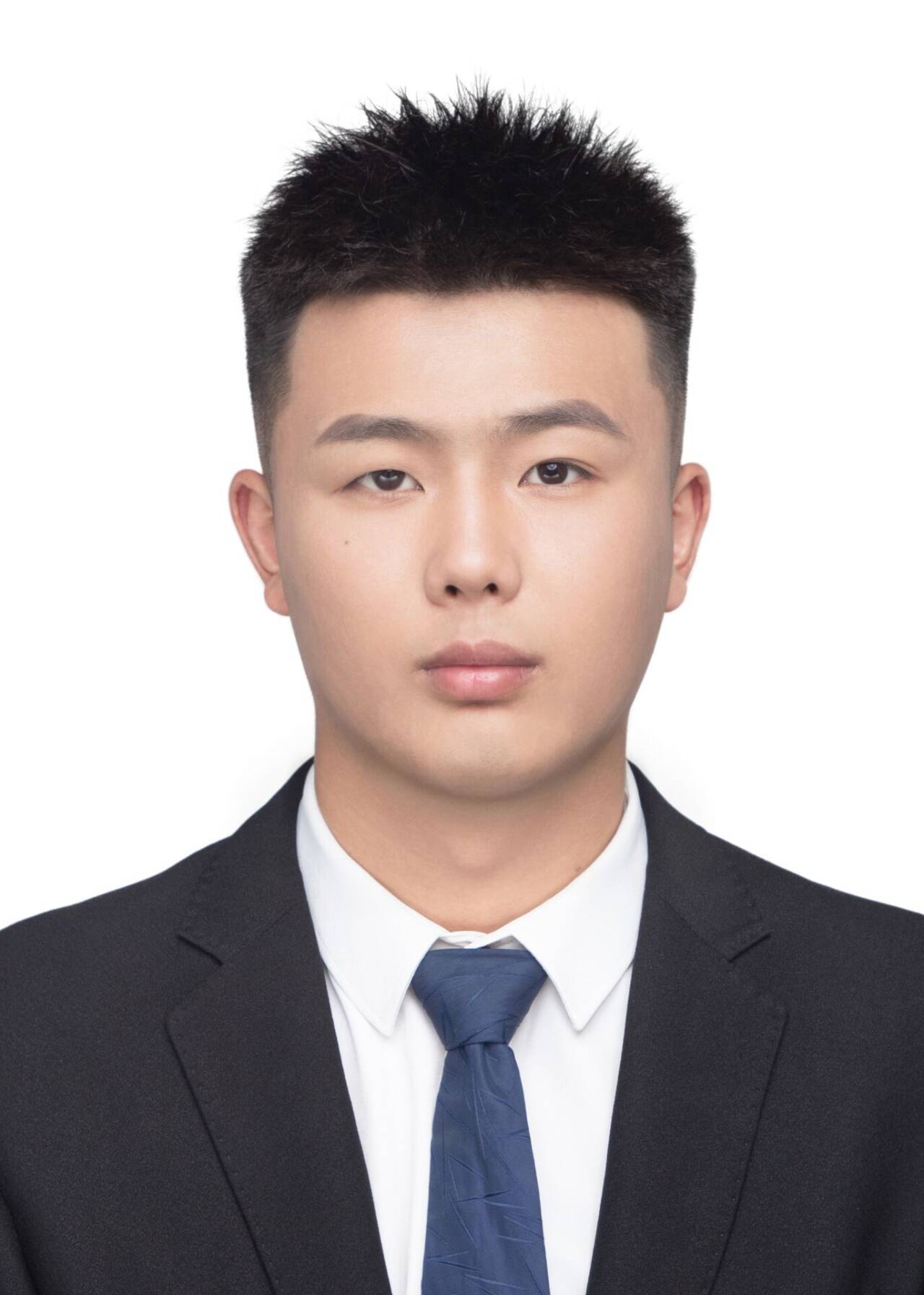}}]{Weizhi Chen}
    \setlength{\parindent}{1em}
    received a bachelor’s degree in engineering, resource exploration engineering from China University of Mining and Technology, Xuzhou, China, in 2023. He is currently pursuing a master's degree with University of Chinese Academy of Sciences, Beijing, China. \par
    His research interests include remote sensing vision-language model, encompassing foundation model and multimodal large language model.
\end{IEEEbiography}

\begin{IEEEbiography}[{\includegraphics[width=1in,height=1.25in,clip,keepaspectratio]{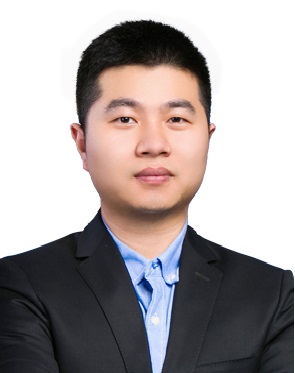}}]{Yupeng Deng}
    \setlength{\parindent}{1em}
    received the Ph.D. degree from the Aerospace Information Research Institute, Chinese Academy of Sciences, Beijing, China, in 2023. \par
    He is currently a Special Assistant Professor with the Aerospace Information Research Institute, CAS. His research interests include computer vision, remote sensing intelligent mapping and change detection. 
\end{IEEEbiography}

\begin{IEEEbiography}[{\includegraphics[width=1in,height=1.25in,clip,keepaspectratio]{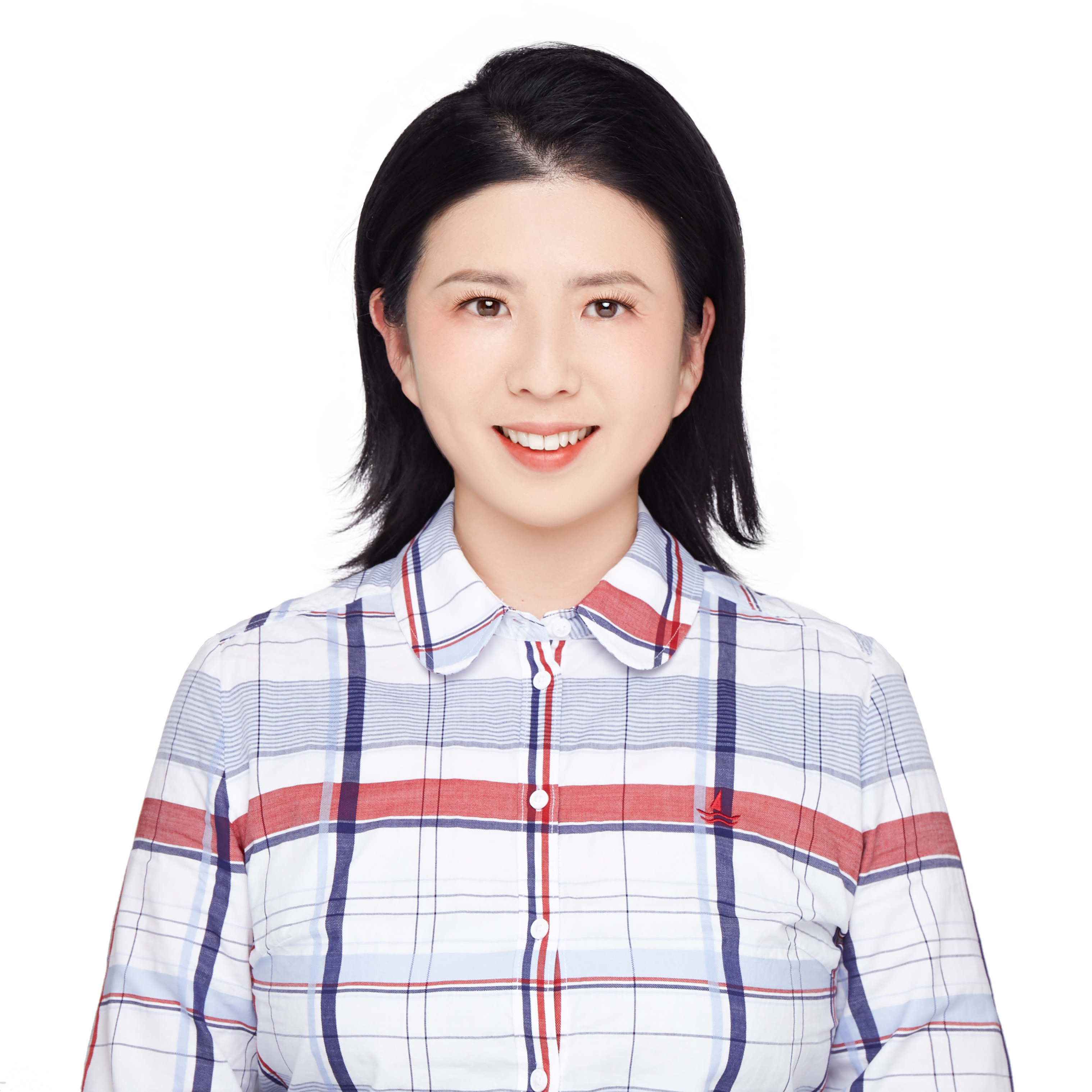}}]{Wei Jin}
    \setlength{\parindent}{1em}
    received her master's degree from the PLA Information Engineering University in 2008. She is currently working as an engineer in PLA Unit 32021. Her research interests include data management, portal website operation and maintenance, etc., with a focus on open-source information and website operation and maintenance applications.
\end{IEEEbiography}

\begin{IEEEbiography}[{\includegraphics[width=1in,height=1.25in,clip,keepaspectratio]{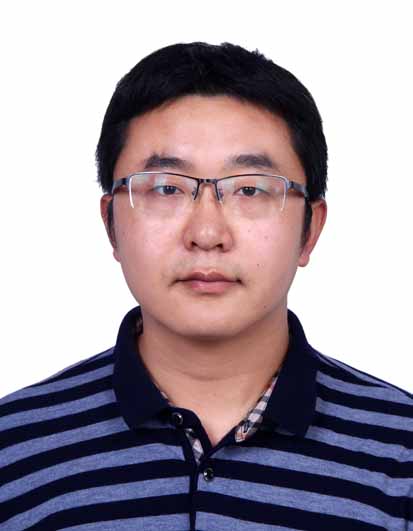}}]{Jingbo Chen}
    \setlength{\parindent}{1em}
    received the Ph.D. degree in cartography and geographic information systems from the Institute of Remote Sensing Applications, Chinese Academy of Sciences, Beijing, China, in 2011. \par
    He is currently a Professor with the Aerospace Information Research Institute, Chinese Academy of Sciences. His research interests cover intelligent remote sensing analysis, integrated application of communication, navigation, and remote sensing.
\end{IEEEbiography}

\begin{IEEEbiography}[{\includegraphics[width=1in,height=1.25in,clip,keepaspectratio]{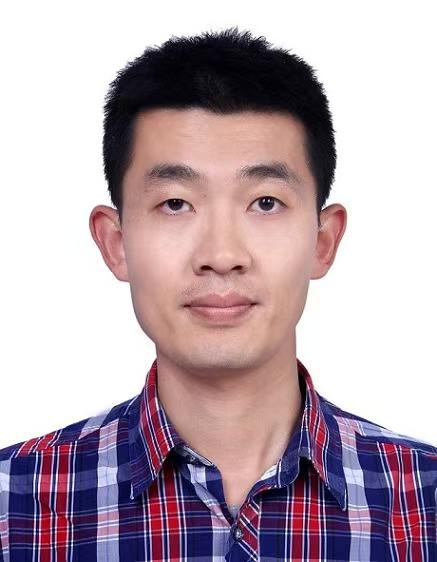}}]{Jiansheng Chen}
    \setlength{\parindent}{1em}
    received the Ph.D. degree in signal and information processing from Institute of remote sensing applications, Chinese Academy of Sciences, Beijing, China, in 2012. \par
    He is currently working with Aerospace Information Research Institute, Chinese Academy of Sciences. His research interests include remote sensing intelligent interpretation and big data governance.
\end{IEEEbiography}

\begin{IEEEbiography}[{\includegraphics[width=1in,height=1.25in,clip,keepaspectratio]{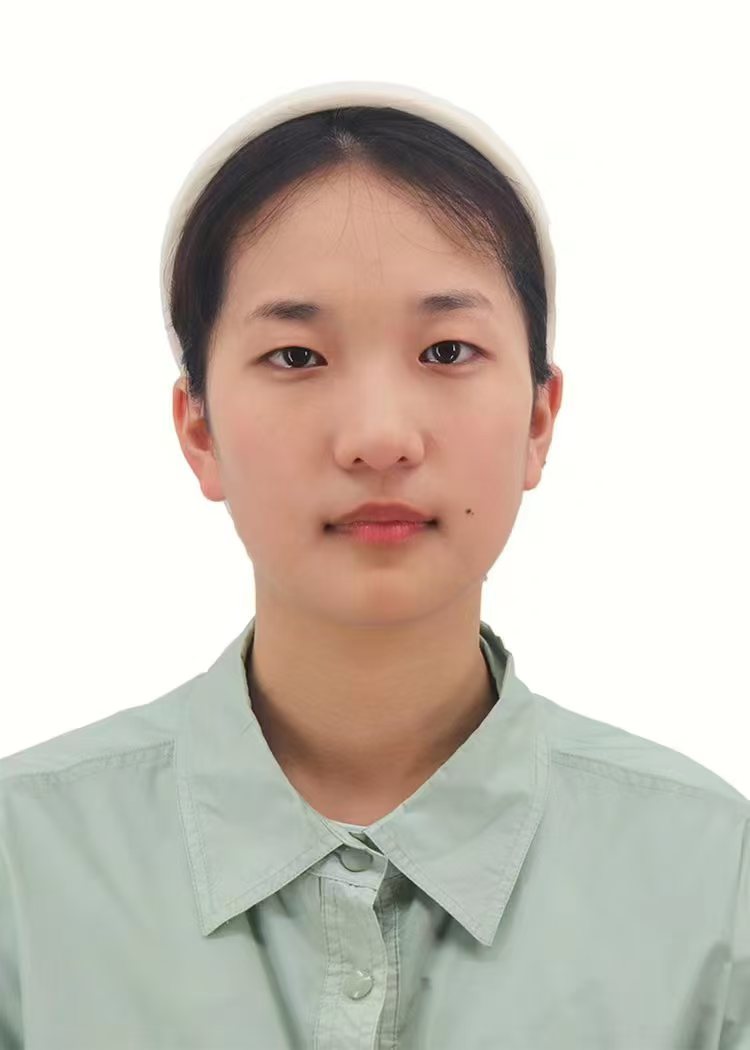}}]{Yuman Feng}
    is currently an undergraduate student majoring in Cybersecurity and Law Enforcement at the School of Information Network Security, People's Public Security University of China. During the holidays, she interned at the Institute of Air and Space Information, Chinese Academy of Sciences. Her research interests include mathematical modeling, artificial intelligence, and network attack and defense.
\end{IEEEbiography}

\begin{IEEEbiography}[{\includegraphics[width=1in,height=1.25in,clip,keepaspectratio]{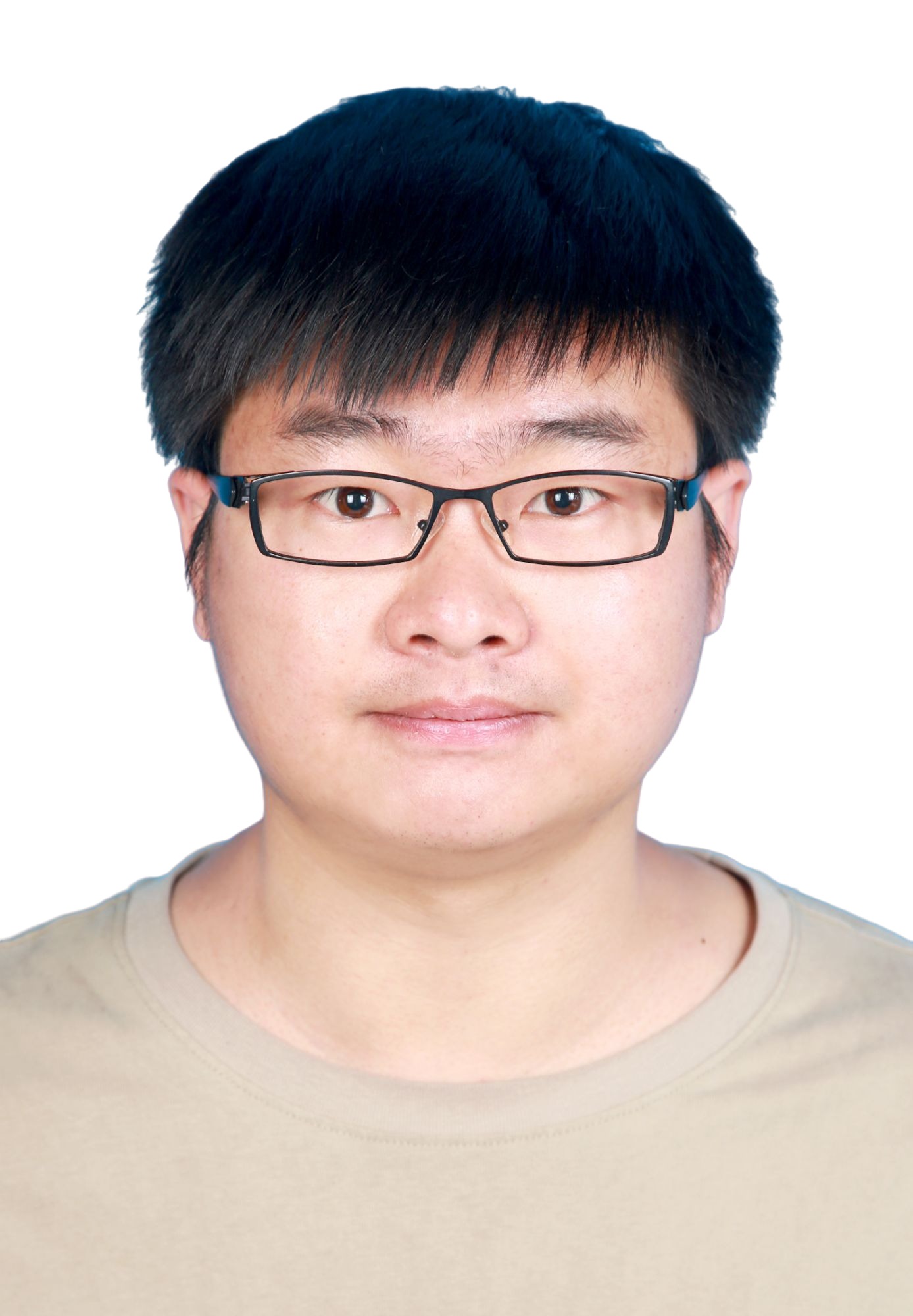}}]{Zhihao Xi}
    \setlength{\parindent}{1em}
    received the Ph.D. degree from the Aerospace Information Research Institute, Chinese Academy of Sciences, Beijing, China, in 2024. \par
    He is currently an Assistant Professor with the Aerospace Information Research Institute, CAS. His research interests include computer vision, domain adaptation, and remote sensing image interpretation.
\end{IEEEbiography}

\begin{IEEEbiography}[{\includegraphics[width=1in,height=1.25in,clip,keepaspectratio]{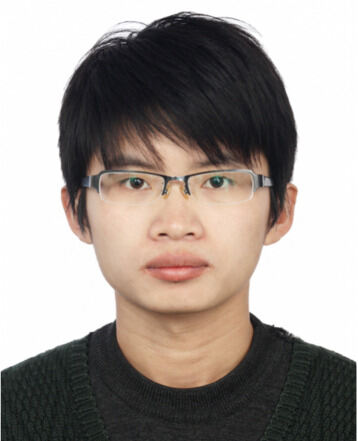}}]{Diyou Liu}
    \setlength{\parindent}{1em}
    received the Ph.D degree in Agricultural Information Technology, College of Land Science and Technology, China Agricultural University in 2021. \par
    He is currently a Special Research Assistant at Aerospace Information Research Institute, Chinese Academy of Sciences, Beijing, China. His research focuses on the production of cartographic-level vector element data using intelligent interpretation methods from high-resolution remote sensing imagery.
\end{IEEEbiography}

\begin{IEEEbiography}[{\includegraphics[width=1in,height=1.25in,clip,keepaspectratio]{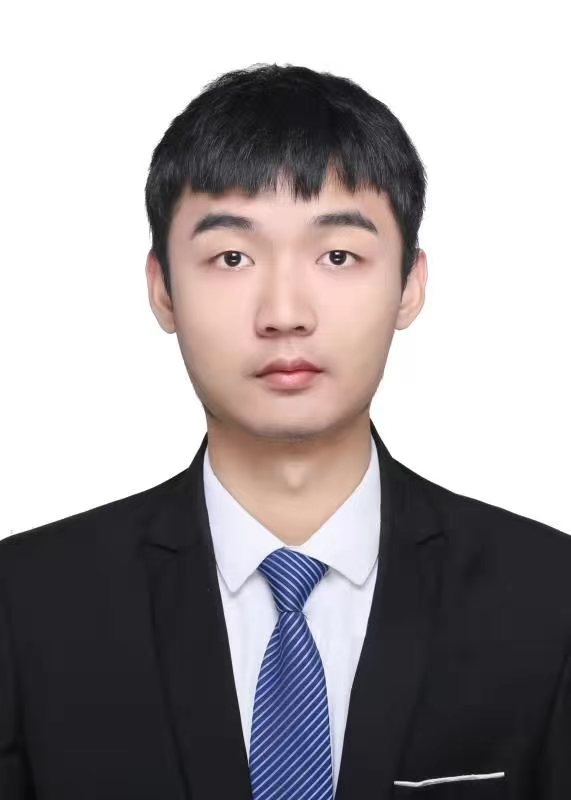}}]{Kai Li}
    \setlength{\parindent}{1em}
    received a bachelor’s degree in engineering, spatial information and digital technology from University of Electronic Science and Technology of China, Chengdu, China, in 2021. He is currently pursuing a PhD degree with University of Chinese Academy of Sciences, Beijing, China. \par
    His research interests include remote sensing, computer vision, and machine learning.
\end{IEEEbiography}

\begin{IEEEbiography}[{\includegraphics[width=1in,height=1.25in,clip,keepaspectratio]{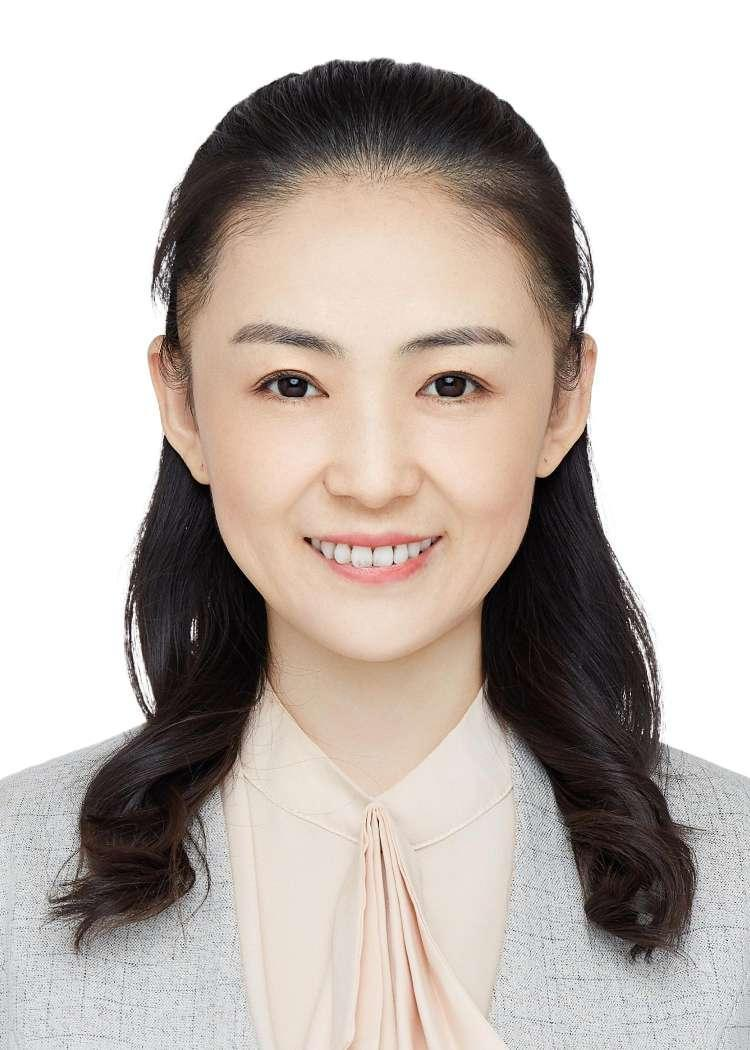}}]{Yu Meng}
    \setlength{\parindent}{1em}
    received the Ph.D. degree in signal and information processing from the Institute of Remote Sensing Applications, Chinese Academy of Sciences, Beijing, China, in 2008. \par
    She is currently a Professor at the Aerospace Information Research Institute, Chinese Academy of Sciences. Her research interests include intelligent interpretation of remote sensing images, remote sensing time-series signal processing, cross-domain intelligent data processing and big spatial-temporal data application. \par
    Dr Meng serves as an editor and board member of the National Remote Sensing Bulletin, Journal of Image and Graphics.
\end{IEEEbiography}

\vspace{11pt}

\end{document}